\documentclass{article}

\usepackage[final]{corl_2026} 

\usepackage{microtype}
\usepackage{graphicx}
\usepackage{wrapfig}
\usepackage{subcaption}
\usepackage{tabularx}
\usepackage{booktabs} 

\usepackage{hyperref}
\usepackage{enumitem}

\usepackage{amsmath}
\usepackage{amssymb}
\usepackage{mathtools}
\usepackage{amsthm}

\usepackage[nohypertypes={main,acronym}]{glossaries}
\usepackage{multirow}

\usepackage[capitalize,noabbrev]{cleveref}

\usepackage[bottom]{footmisc}

\usepackage{tcolorbox}
\tcbuselibrary{skins}
\usepackage{listings}

\newtcolorbox{promptbox}[1][]{
    colback=gray!5,
    colframe=gray!60,
    arc=3mm,
    boxrule=0.5pt,
    left=10pt, right=10pt, top=8pt, bottom=8pt,
    fonttitle=\bfseries,
    title=#1,
    coltitle=black,
    colbacktitle=gray!30,
    enhanced,
    attach title to upper={\par\smallskip},
    toptitle=6pt,
    bottomtitle=6pt,
}

\theoremstyle{plain}

\theoremstyle{definition}

\theoremstyle{remark}

\title{Adaptive World-Models for Anomaly Recognition (AWARE): A Framework for Robot Dynamics Prediction and Introspection}

\author{
\textbf{Luke Beddow} \quad
\textbf{George Mavroghenis} \quad
\textbf{Rodrigo Chacon Quesada} \quad
\textbf{Kamil Dreczkowski} \\
\textbf{Cong Sun} \quad
\textbf{Jiankai Wang} \quad
\textbf{Oscar Kwong-Fai Pang} \quad
\textbf{Antoine Cully} \\[0.5em]
BOLD, Imperial College London \\[0.5em]
\textbf{Project page:} \url{https://adaptive-intelligent-robotics.github.io/aware/} \\
\textbf{Code:} \url{https://github.com/adaptive-intelligent-robotics/aware}
}

\begin{document}
\maketitle


\newacronym{ale}{ALE}{Adaptive Latent Estimator}
\newacronym{auprc}{AUPRC}{Area Under the Precision-Recall Curve}
\newacronym{auroc}{AUROC}{Area Under the Receiver Operating Characteristic}
\newacronym{aware}{AWARE}{Adaptive World-model for Anomaly REcognition}
\newacronym{dof}{DoF}{Degrees of Freedom}
\newacronym{hdi}{HDI}{Highest Density Interval}
\newacronym{mujoco}{MuJoCo}{ Multi-Joint dynamics with Contact.}
\newacronym{pe}{PE}{Prediction Error}
\newacronym{physmop}{PhysMoP}{Physics-based Motion Prediction}
\newacronym{roa}{ROA}{Regularised Online Adaptation}
\newacronym{rope}{ROPE}{Region of Practical Equivalence}
\newacronym{rssm}{RSSM}{Recurrent State-Space Model}
\newacronym[plural=World Models]{wm}{WM}{World Model}
\newacronym{ap}{AP}{Average Precision}
\newacronym{rwm}{RWM}{Robotic World Model}
\newacronym{ROA_dec}{ROA\_dec}{ROA+Decoder}
\newacronym{ROA_dec_noLB}{ROA\_dec\_noLB}{ROA + Decoder + No Latent Bottleneck}
\newacronym{ROA_dec_noReg}{ROA\_dec\_noReg}{ROA + Decoder + No  Regularization}
\newacronym{ROA_noLB}{ROA\_noLB}{ROA + No Latent Bottleneck}
\newacronym{ROA_noLB_noReg}{ROA\_noLB\_noReg}{ROA + No Latent Bottleneck + No Regularization}
\newacronym{ROA_noReg}{ROA\_noReg}{ROA + No Regularization}
\newacronym{T_AWARE}{T\_AWARE}{Transformer AWARE}
\newacronym{T_ROA}{T\_ROA}{Transformer ROA}

\begin{abstract}

Unrecognized anomalies in critical infrastructure can rapidly cascade into catastrophic physical or operational failures. Current anomaly detection approaches often rely on direct access to the monitored system by utilizing onboard sensors, such as motor encoders, to detect anomalous states. However, legacy hardware often lacks these sensors, or strict security protocols may prohibit access altogether. We therefore present \acrfull{aware}, a world model framework that is capable of anomaly detection using only external camera monitoring via system dynamics estimation and state prediction. Evaluated across $240$ real-world robotic crane trajectories, including payload collisions and motor degradations, AWARE demonstrates a $26\%$ improvement in anomaly classification over our strongest baseline  under noisy, camera-only state estimation. Furthermore, the inferred dynamics parameter can be decoded to identify the source of faults, localizing the degrading motor with $88\%$ accuracy. 

\end{abstract}


\begin{figure}[h]
    \centering
    \includegraphics[width=0.98\linewidth]{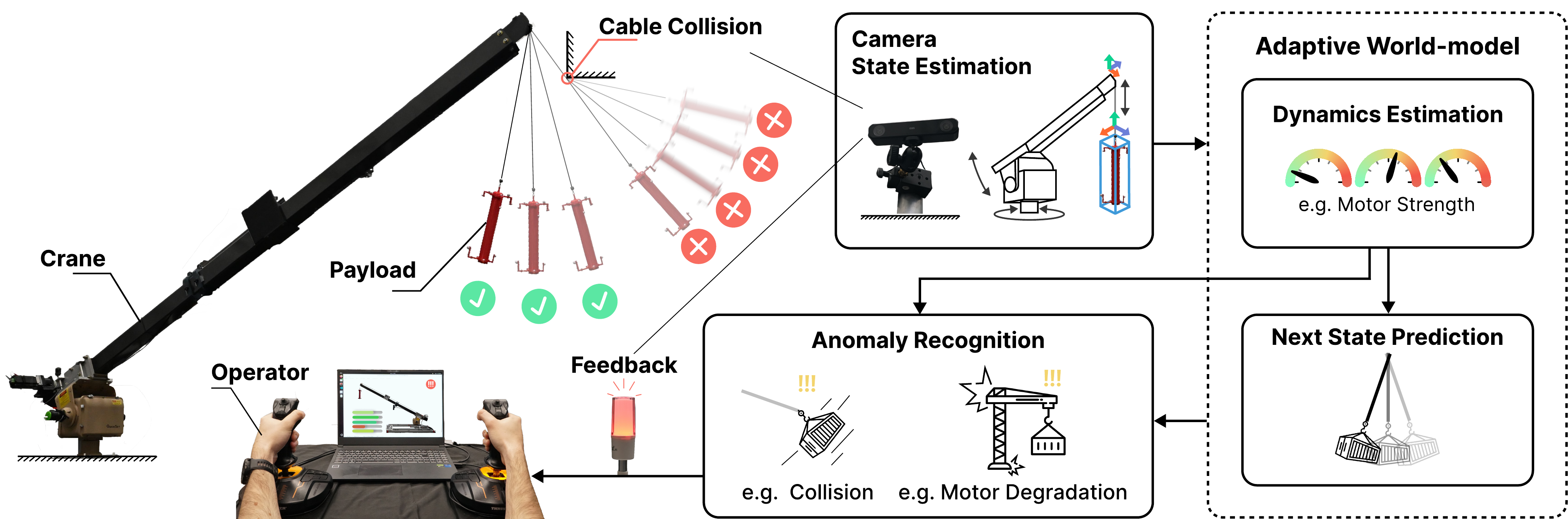}
    \caption{\small {System and model overview, showing human-operated robotic crane used for anomaly detection experiments. \acrshort{aware} operates as follows: 1) camera-based state estimation; 2) estimation of system hidden dynamics parameters; 3) prediction of future trajectory; 4) anomaly recognition via examination of both the estimated dynamics parameters and the trajectory prediction error.}}
    \label{fig:teaser}
\end{figure}

\section{Introduction}

Critical infrastructure, such as ports, power plants, and factories, heavily relies on robotic platforms to perform tasks~\cite{nandakumar2024anomaly}. Disruption to these systems, such as through unauthorized operation or cyberattacks, can be physically dangerous to nearby workers, and cause significant economic disruption. Consequently, monitoring these systems for anomalous behavior is crucial to ensure safe and continuous nominal operation.

Traditional learning-based anomaly detection approaches focus either on spatial reconstruction methods for detecting unexpected features in high-dimensional state data (e.g. via variational auto-encoders), or temporal sequence methods for forecasting and measuring divergence from predicted states (e.g. via dynamics models) ~\cite{bogdoll2022anomaly, Domberg2025WMforAD}. However, these methods require precise estimation of the current robot state, typically via access to onboard sensors, such as motor encoders~\cite{Lu2022anomaly}. Many infrastructure applications prevent this low-level access due to lacking the required sensors or because of strict security protocols. Such applications, necessitate a physical 'air-gap' between the state estimation system and the observed hardware, such as through an external camera. This form of external passive state estimation injects noise throughout the entire anomaly detection pipeline and prevents sufficient system identification, drastically degrading the performance of learned models.

Foundation model-based approaches, (e.g. LLMs, VLMs) offer a promising alternative ~\cite{Ahn2026WACV, SinhaElhafsiEtAl2024} . Instead of relying on precise state data, they leverage a more general semantic understanding of the scene (i.e. directly from vision or text descriptions) to detect  anomalies. However this comes with a clear trade-off: sacrificing a precise understanding of temporal dynamics for semantic generalizability, which may prohibit detection of subtle, mechanics-based anomalies.

To bridge this gap, we introduce \textbf{Adaptive World-Models for Anomaly Recognition (AWARE)}: a framework for anomaly detection under high-uncertainty camera-only state estimation. \acrshort{aware} jointly trains a latent dynamics estimator and world model. The estimator provides a latent representation of the robot's intrinsic parameters, such as motor gain or mass, while the world model uses this representation, along with an intended action, to produce an accurate prediction of the next robot state under the assumed dynamics. \acrshort{aware} directly addresses the three prior challenges of passive monitoring by introducing noise resilience via a novel dual-signal architecture, relaxing system identification requirements through domain randomization and world model dynamics conditioning, and providing explainable anomaly detection by identifying the fault source. \autoref{fig:teaser} shows the AWARE system overview.

The key contributions of this work are:
\begin{itemize}[topsep=0pt, partopsep=0pt, noitemsep]
    \item \textbf{AWARE framework}. An adaptive world model conditioned on a latent dynamics estimator designed for anomaly detection under high noise state estimation. AWARE is trained fully in simulation using domain randomization, without requiring perfect system identification or explicit priors on anomaly types.  
    \item \textbf{Robust detection and identification}. We demonstrate that a novel dual-signal architecture of world model prediction error and latent dynamics deviation improves the detection of in and out-of-distribution anomalies on a challenging real world platform. AWARE shows a modest, but statistically significant, $3\%$ improvement over the next best baseline under low-noise state estimation, and a significant $26\%$ performance increase while identifying the source of in-distribution anomalies in $88\%$ of cases under high-noise state estimation.
    \item \textbf{Rigorous real-world experimental analysis}. We evaluate AWARE over $240$ real world robotic crane trajectories driven by $5$ different human operators. We perform a detailed analysis and ablation of the AWARE components and compare with a comprehensive selection of baselines including simulators, world models, foundation models, dynamics estimators and established anomaly detection approaches. 
\end{itemize}

\section{Related Work}

Anomaly detection in robotics involves discerning unexpected states or behaviors~\cite{nandakumar2024anomaly}. Statistical approaches~\cite{Schnell2020robot} and reconstruction techniques~\cite{Nehme2021anomaly} score observations by their likelihood under a reference distribution. While effective for single-state anomalies, infrastructure monitoring must additionally capture temporal dynamics changes and distinguish anomalies from valid operator actions. Recent spatiotemporal methods address this for driving, cyber-attacks, and navigation failures~\cite{Ammar2026transformerAD,kang2025real,Fang2022traffic,Zekry2021anomaly,Ji2022proactive,rahman2024large}, but assume labeled anomaly data that is unavailable in semi-supervised infrastructure settings~\cite{Brockmann2024ADinRobotics}. Foundation-model approaches offer zero-shot semantic understanding to derive anomaly through visual or language prompts~\cite{Ahn2026WACV,SinhaElhafsiEtAl2024,agia2025sentinel,liu2025siriusfleet}. However, these methods reason over semantic scene features and are not sensitive to subtle perturbations in system dynamics (e.g. gradual motor degradation or shifts in inertial properties) that weakly manifest in visual appearance. Predictive temporal models instead learn nominal dynamics and flag deviations~\cite{Hundman2018detecting,Lu2022anomaly}, but prior work assumes dense onboard sensing. To handle sensing uncertainty without onboard access, robotics uses estimator-based consistency statistics~\cite{darabi2024starnet,Hafez2024integrity} and distribution-free calibration~\cite{Luo2024conformal,xu2025failuredetection}. Unlike prior work, AWARE addresses passive monitoring with no onboard sensing, combining a dynamics-aware world model with latent parameter estimation to detect anomalies that are invisible to appearance-based methods yet critical for safe operation.

World models predict future states given current state and action, and are widely used in model-based RL~\cite{Hafner2025mastering,hafner2025training,wu2022daydreamer,grillotti2025tabula,bar2025navigationworldmodels}. The Dreamer series established the \acrfull{rssm}~\cite{hafner2019pixels} as a prominent backbone; \acrfull{rwm}~\cite{li2025robotic} extended this to output task-relevant quantities alongside state predictions. Existing applications to anomaly detection~\cite{bogdoll2023exploring,Domberg2025WMforAD} demonstrate success through state prediction and reconstruction. AWARE goes further through the latent dynamics estimator, providing an explicit prior on the assumed dynamics for the world model and an additional signal, the latent deviation, for more robust anomaly detection. Estimation of such dynamics parameters via domain randomization in simulation has been demonstrated in many RL control policies~\cite{kumar2021rma,sanghvi2025occam,cheng2024parkour}. \acrfull{roa}~\cite{fu2023deep} trains a teacher--student estimator whose latent dynamics predictions condition a control policy. AWARE adapts this for anomaly detection by learning a teacher encoder through world model prediction gradients and a more informative representation by removing teacher regularization and latent bottleneck whilst including a decoder for introspection.

\section{Adaptive World-Models for Anomaly Recognition (AWARE)}\label{sec: model}


\begin{figure}
\centering
\includegraphics[width=\linewidth]{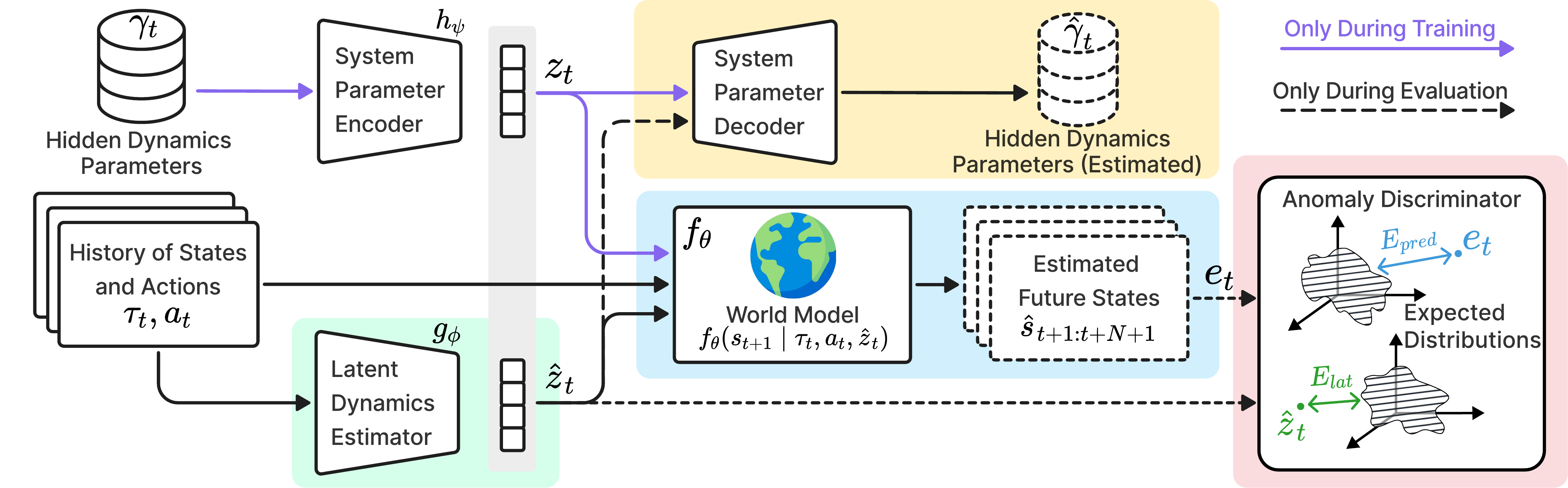}
\caption{\small {\acrshort{aware} framework. Our model consists of two key components. First, a \textbf{Latent Dynamics Estimator} (green) reconstructs a latent encoding of system dynamics ($\hat{z}_{t}$) using a history of state-action pairs. Second, a \textbf{World Model} (blue) predicts the next state, given this encoding, the same history, and the current action. This integrated framework enables \textbf{introspection} (yellow) and \textbf{anomaly detection} (red) by comparing dynamics estimations and state prediction errors ($e_t$) against their expected distributions under normal operation.}}

\label{fig:systemarch}
\end{figure}

We present \textbf{AWARE}, an adaptive world model framework, in \autoref{fig:systemarch}. A latent dynamics estimator predicts an encoding of the system dynamics, which conditions a world model to predict the next state. The latent dynamics estimation and next state prediction error are jointly used to recognize anomalies by comparing them with their expected distribution under normal operation.

\subsection{Problem Formulation}\label{subsec:problem-formulation}

We model the environment as a parametrized stochastic dynamical system, defined by the tuple $(\mathcal{S}, \mathcal{A}, \mathcal{P}, \mathcal{U})$, where  $\mathcal{S}\subseteq\mathbb{R}^{n_s}$ and $\mathcal{A}\subseteq\mathbb{R}^{n_a}$ denote continuous state and action spaces, and $\mathcal{U}\subseteq \mathbb{R}^{n_u}$ is the dynamics context. Each context $u = (\gamma, \zeta) \in \mathcal{U}$ comprises modeled dynamics parameters $\gamma$, such as motor characteristics, and unmodeled dynamics $\zeta$, such as disturbance forces and unexpected contacts. The probabilistic transition function is defined as  $\mathcal{P}:\mathcal{S}\times\mathcal{A}\times\mathcal{U}\rightarrow\Delta(\mathcal{S})$, and transitions at timestep $t$ are sampled according to $s_{t+1}\sim\mathcal{P}(\cdot\mid s_t,a_t,\gamma_t,\zeta_t)$.

We define anomalies as when the dynamics context changes from an initial value $u_0 = (\gamma_0, \zeta_0)$ to a new value $u_t = (\gamma_t, \zeta_t)\neq u_0$. Changes in $\gamma_t$ are detected by estimating it via latent representation $\hat{z}_t$. Since $\zeta_t$ cannot be estimated directly, changes in it are instead detected through errors in predicting the next state $\hat{s}_{t+1}$.


\acrshort{aware} detects anomalies by taking as input an $H$-step history of states and actions, $\tau_t = (s_{t-H:t}, a_{t-H:t-1})$, along with the current action $a_t$ and predicting two quantities: $\hat{z}_t$ and $\hat{s}_{t+1}$. Each prediction is converted to an anomaly score based on its error against a reference distribution characterizing normal operation. \autoref{fig:systemarch} illustrates the \acrshort{aware} architecture.

\subsection{Dynamics Parameter Estimation}

Identifying true robotic system dynamics requires complex system identification~\cite{lee2023robot}, which is particularly challenging when only noisy state estimates are available. \acrshort{aware} instead learns to estimate a representation of the dynamics from a history of states and actions using a method based on \textit{Regularized Online Adaptation} (\acrshort{roa}) ~\cite{fu2023deep}.

Training occurs in simulation, by varying a predefined set of modeled dynamics parameters, $\gamma$, such as motor strength or body inertia. Such parameters may be unknown on the real system, but can typically be bounded within a range $\gamma \in [\gamma_{min}, \gamma_{max}]$, with a wider range reflecting greater uncertainty.

During training, $\gamma$ is randomly sampled and applied to the physics simulation. Following the teacher-student method of \acrshort{roa}, a teacher encoder $h_{\psi}$ maps the true parameters to a latent representation $z_t=h_\psi(\gamma_t)$, shaped by the downstream world model. A student latent dynamics estimator $g_{\phi}$ then learns to predict this representation from a history of states and actions $\hat{z}_t = g_\phi(s_{t-H:t}, a_{t-H:t})$ by minimizing the following loss:  \begin{equation}
    \mathcal{L}_\phi=||g_\phi(s_{t-H:t}, a_{t-H:t})- h_\psi(\gamma_t)||^2
\end{equation}
We make three key changes to adapt \acrshort{roa} for anomaly detection. First, the downstream model that shapes $z$ is a world model rather than a policy, meaning $z$ is optimized for dynamics prediction rather than control (Sec.~\ref{subsec:method-world-model}). Second, we add a System Parameter Decoder that recovers $\gamma$ from $\hat{z}$. This supports introspection into which aspect of the dynamics has changed. Third, we remove both compression and regularization of $z$ used in the original \acrshort{roa} method, to preserve as much information as possible in the latent representation. 

\subsection{World Model State Prediction} \label{subsec:method-world-model}

We train a world model $f_{\theta}$ to approximate the true transition function $\mathcal{P}$. Conditioned on the history $\tau_t$, the current action $a_t$ and the latent dynamics $z$ ($\hat{z}_t$ at deployment), it outputs a Gaussian over the next state with diagonal covariance, whose mean is the prediction $\hat{s}_{t+1}$:
\begin{equation}
    f_\theta(s_{t+1} \mid \tau_t, a_t, z) = \mathcal{N}\!\big(\hat{s}_{t+1},\, \operatorname{diag}(\hat{\sigma}_{t+1}^2)\big).
\end{equation}
To show that \acrshort{aware} is architecture-agnostic, we evaluate two backbones: a Transformer, which attends over $\tau_t$ directly, and an \acrshort{rssm} ~\cite{hafner2019pixels} , which summarizes the history in a recurrent hidden state $h_t$. Both are trained by minimizing the negative log-likelihood $\mathcal{L}_\theta = -\log f_\theta(s_{t+1} \mid \tau_t, a_t, z)$.

\subsection{Anomaly Decision} \label{subsec:anomaly-decision}

An anomaly score is derived from the estimated latent dynamics parameters $\hat{z}_t$ as well as the prediction of the next state $\hat{s}_{t+1}$. The latter, is converted into an error $e_t$ against the observed next state $s_{t+1}$ and averaged over a prediction window of $N$ time-steps. The prediction error is normalized by the predicted standard deviation $\hat{\sigma}$, reducing the contribution of prediction errors to the anomaly score when the world model expresses higher uncertainty:
\begin{equation}\label{equation:anomaly-pred}
    e_t = \frac{1}{N}\sum_{k=t-N}^{t-1} \lvert s_{k+1} - \hat{s}_{k+1} \rvert \oslash \hat{\sigma}_{k+1}.
\end{equation}
During initialization, a set of reference trajectories is collected to characterize the distribution of nominal system operation with respect to the dynamics parameters $\hat{z}_0$ and model error $e_0$. The reference trajectories may be collected prior to initialization as long as the robot context dynamics remain fixed up to the current operating point.  

At test-time, the Mahalanobis distance \cite{mahalanobis1936generalized} between the state error, dynamics latent, and their respective reference distributions, is computed to give two anomaly scores $E_{lat}$ and $E_{pred}$. $E_{lat}$ arises from the latent dynamics estimation $\hat z_t$ which is intended to detect changes in the modeled dynamics $\gamma$. $E_{pred}$ derives from the prediction error $e_t$ and aims to detect changes in unmodeled dynamics $\zeta$. Additionally, by conditioning the world model on assumed nominal dynamics $\hat{z}_0$ (as opposed to $\hat{z}_t$) we provide $E_{pred}$ with additional sensitivity to changes in $\gamma$. Anomaly decisions use two independent thresholds for $E_{lat}$ and $E_{pred}$, where each score represents the distance in standard deviations that an incoming data point is from the reference distribution. Thresholds are calibrated during nominal operation considering specific operational sensitivity requirements.   

\section{Experimental Evaluation}

\textbf{Experimental Procedure and Task Design. }The models were trained using data generated from a MuJoCo simulation of the crane. The simulation was calibrated to the real platform through a grid search over parameters $\gamma$. We then generated data using an exploratory policy with uniform domain randomization centered over the ground-truth values. This randomization therefore removes the requirement for perfect identification of $\gamma$. Appendix~\ref{app:model-hyperparameters} details the parameterized values and domain randomization ranges. 

\begin{figure}[t]
\centering
\includegraphics[width=\linewidth]{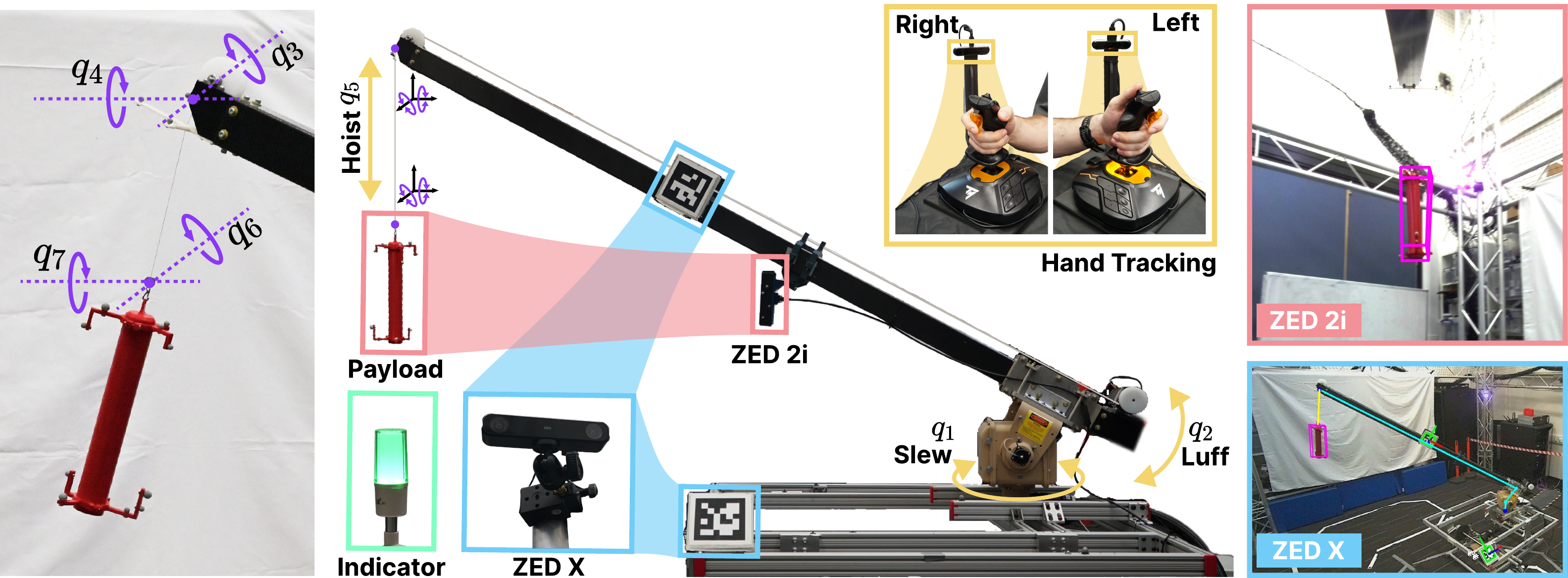}
\caption{
Deployment Method \textbf{(1) Crane platform}: The system is modeled by three actuated joints: slew ($q_1$), luff ($q_2$), and hoist ($q_5$). The under-actuated dynamics are captured by a double pendulum model: two angles ($q_3$, $q_4$) defining the movement at the hoist connection point (cable swing), and two angles ($q_6$, $q_7$) defining the relative orientation of the payload with respect to the cable. 
\textbf{(2) Camera-Based State Estimation}: Operator actions are estimated by two hand-tracking cameras. Crane states $q_1$ and $q_2$ are derived from ArUco marker poses using a ZED X camera. The remaining states ($q_3$–$q_7$) are estimated by a boom-mounted ZED 2i camera utilizing a Convolutional Neural Network (CNN).
\textbf{(3) Operator Interface}: A multi-color indicator lamp provides real-time visual feedback, signaling nominal operation (green light) or system alerts (red light).
}
\vspace{-0.4cm}
\label{fig:methods}
\end{figure}

Our robotic crane is an under-actuated system with three actuated joints, slew ($q_1$, revolute), luff ($q_2$, revolute), hoist ($q_5$, prismatic), and four free joints, formed by the suspended payload and the attached cable, with ($q_3, q_4$) at the boom captures pendulum swing and ($q_6, q_7$) at the payload captures double-pendulum effects when the payload is not aligned with the cable (~\autoref{fig:methods}). 

We collected human-in-the-loop crane trajectories with one of five human operators continuously moving the payload through three randomly sampled waypoints. Two state-estimation pipelines ran in parallel to produce the robot state $s_t$ and operator action $a_t$ at each timestep: one using high-quality motion capture, and the other using RGB cameras. To make sure AWARE is evaluated on system anomalies other than perception failures (e.g. sensor blockage, markers or
payload leaving the field of view), we screened the dataset during collection to ensure such episodes are absent. In total, we collected 240 trajectories, comprising 80 nominal runs and 160 anomalous runs. 

We consider two anomaly families. \textit{Payload collision} (40 trajectories) introduces an unmodeled disturbance $\zeta$ via a single impact to the payload or cable during the anomaly window. \textit{Motor degradation} (120 trajectories) scales down the executed velocity of one of the three motors, representing anomalies in modeled dynamics $\gamma$. For each motor, 40 trajectories are evenly split across $\{75, 50, 25, 0\}\%$ of nominal velocity. Since training randomizes motor strength within $50 - 150\%$, $75\%$ and $50\%$ are in-distribution while $25\%$ and $0\%$ are out-of-distribution. We compute anomaly scores as a continuous (per-timestep) classification signal and adopt a trajectory-level criterion following prior work~\cite{Hundman2018detecting}: a nominal trajectory must contain no anomaly decisions, while an anomalous trajectory must trigger at least one anomaly decision within the anomaly window. 

\textbf{State Estimation. }Robot joint positions ($q_1, \dots, q_7$) and operator actions ($a_1, a_2, a_3$), were measured for every trajectory using two different state estimation pipelines, all at $20$ Hz.

Firstly, \textit{Motion Capture Quality Estimation} captures high quality data. This case uses motion tracking cameras to measure the payload position relative to the crane boom, and extract the underactuated joint angles and hoist length ($q_3, \dots, q_7$). The remaining joints are measured using proprioception, motor encoders for slew and luff joints ($q_1, q_2$), and the exact joystick position for operator actions ($a_1, a_2, a_3$). The resultant state estimation approaches the ground truth, but noise and measurable calibration error arising from the motion capture pipeline reduce data quality.

Secondly, \textit{Camera Only Estimation}: a fully passive state estimation pipeline using only exteroception. This data contains considerably higher levels of noise, and was motivated by the application of infrastructure monitoring, which can require entirely exteroceptive monitoring because proprioceptive sensors may be blocked due to security requirements or constraints of legacy equipment.

Four cameras estimate state and action: a ZED X for slew and luff joints ($q_1,q_2$) via ArUco markers, a boom-mounted ZED 2i with a CNN (EfficientNet-B2 backbone~\cite{tan2019efficientnet}) for payload joints ($q_3, \dots, q_7$), and two Leap Motion cameras for operator actions via an MLP. Joint velocities are derived by finite difference after outlier filtering.

\textbf{Metrics. }We define the anomaly detection experiment as trajectory level classification, derived from per-timestep anomaly decisions. This results in a binary label for each of the $240$ trajectories in the experiment. Performance was evaluated using \acrfull{ap}, a threshold-independent measure, and F1 Score. To standardize dual-threshold comparison, we fix the threshold grid at $32\times32$ and use interpolated \acrshort{ap}, which constructs a monotonic Pareto front treating single and dual threshold methods identically. Note that a random signal achieves $0.17$ \acrshort{ap}. Generalization is assessed via F1 score using stratified 5-fold cross validation, with thresholds selected on validation folds to prevents dual threshold overfitting. All metrics were averaged over $10$ repeats, each with a random sample of $10$ anomaly free trajectories as the reference distribution for calculating the Mahalanobis distance. Statistical significance was assessed using a two-tailed Wilcoxon signed-rank test over $10$ random model seeds for both \acrshort{ap} and F1 score. Given the two metrics and two state estimation conditions, we applied the Bonferroni correction for an adjusted significance level of $\alpha=0.05/4=0.0125$.

\textbf{Model Baselines. }We demonstrate that AWARE is model agnostic by evaluating with both a Transformer and \acrshort{rssm} world model, in line with previous work~\cite{li2025robotic, alonso2024diffusion, hansen2024tdmpc}. We also evaluate both as stand-alone world model baselines (without the full \acrshort{aware} method) alongside the fine-tuned MuJoCo model from Appendix~\ref{app:model-hyperparameters}.

We select two non-temporal, non-model based anomaly detection baselines: Isolation Forest~\cite{liu2008isolation} and Statistical Anomaly Detection. The former is an unsupervised tree-based method that aims to isolate anomalous data-points via random partitioning  and the latter measures the Mahalanobis distance of each data-point to a reference distribution. Both work using raw states and actions.
We evaluate two foundation model based methods: AESOP~\cite{SinhaElhafsiEtAl2024} and AnyAnomaly~\cite{Ahn2026WACV}. AESOP generates an embedding from text-descriptive states and actions using a small LLM, computing anomaly signals via the Mahalanobis distance to a reference distribution. AnyAnomaly uses a VLA to infer if an anomaly is occurring in a batch of image frames, requiring the type anomaly to be specified in the prompt. As AnyAnomaly only operates over a short sequence of frames, we only evaluate it on the payload collision anomaly set. See Appendix~\ref{app: baselines} for the implementation details. 

We use the \acrshort{roa}~\cite{fu2023deep} method as a baseline, training a latent estimator with $\gamma$ as the privileged information and identical domain randomization. This allows for latent dynamics anomaly scores. We train it via learning a policy to navigate the crane payload to specific targets in MuJoCo. Finally, we implement RWM~\cite{li2025robotic}, as the closest related world model to our method. We set their model's privileged information to $\gamma$, which whilst outside of the scope of the in-domain learning presented by their method, is valid for architectural comparison. This results in both prediction error and dynamics estimation anomaly scores.

\begin{table}
\begin{small}
\caption{Anomaly recognition results over $240$ real robotic trajectories, averaged over $10$ seeds per trained model. Metrics detail classification performance for identifying anomalies at the correct times within trajectories, under two different passive state estimation schemes. Bold entries are statistically significant improvement over the best non-AWARE model. $^*$AnyAnomaly is only evaluated on the nominal and payload collision set, see Appendix~\ref{app: baselines} for more detail.}
\label{tab:discrim-results}
\centering

\label{tab:my-table}
\begin{tabular}{@{}lcccc@{}}
\toprule
\multirow{2}{*}{Method}                  & \multicolumn{2}{c}{\begin{tabular}[c]{@{}c@{}}Motion Capture Quality\\ State Estimation (Low Noise)\end{tabular}}                  & \multicolumn{2}{c}{\begin{tabular}[c]{@{}c@{}}Camera Only State\\ Estimation (High Noise)\end{tabular}}                            \\ \cmidrule(l){2-5} 
                                         & \begin{tabular}[c]{@{}c@{}}Avg.\ Precision\\ $\pm$ std\end{tabular} & \begin{tabular}[c]{@{}c@{}}F1 Score\\ $\pm$ std\end{tabular} & \begin{tabular}[c]{@{}c@{}}Avg.\ Precision\\ $\pm$ std\end{tabular} & \begin{tabular}[c]{@{}c@{}}F1 Score\\ $\pm$ std\end{tabular} \\ \midrule
Randomly Generated Signal                & 0.17\,$\pm$\,0.03                                                   & 0.26\,$\pm$\,0.03                                            & 0.17\,$\pm$\,0.03                                                   & 0.26\,$\pm$\,0.03                                            \\
Isolation Forest~\cite{liu2008isolation} & 0.29\,$\pm$\,0.00                                                   & 0.36\,$\pm$\,0.01                                            & 0.16\,$\pm$\,0.00                                                   & 0.25\,$\pm$\,0.01                                            \\
Statistical Anomaly Detection            & 0.40\,$\pm$\,0.01                                                   & 0.40\,$\pm$\,0.01                                            & 0.22\,$\pm$\,0.01                                                   & 0.31\,$\pm$\,0.02                                            \\
AESOP~\cite{SinhaElhafsiEtAl2024}        & 0.47\,$\pm$\,0.00                                                   & 0.56\,$\pm$\,0.00                                            & 0.24\,$\pm$\,0.00                                                   & 0.41\,$\pm$\,0.00                                            \\
AnyAnomaly$^*$~\cite{Ahn2026WACV}            & 0.37\,$\pm$\,0.07                                                   & 0.56\,$\pm$\,0.06                                            & 0.37\,$\pm$\,0.07                                                   & 0.56\,$\pm$\,0.06                                            \\
MuJoCo~\cite{todorov2012mujoco}          & 0.77\,$\pm$\,0.00                                                   & 0.77\,$\pm$\,0.00                                            & 0.45\,$\pm$\,0.00                                                   & 0.44\,$\pm$\,0.00                                            \\
RSSM~\cite{Hafner2025mastering}          & 0.86\,$\pm$\,0.01                                                   & 0.80\,$\pm$\,0.01                                            & 0.53\,$\pm$\,0.01                                                   & 0.56\,$\pm$\,0.01                                            \\
Transformer~\cite{vaswani2017attention}  & 0.84\,$\pm$\,0.01                                                   & 0.79\,$\pm$\,0.01                                            & 0.41\,$\pm$\,0.02                                                   & 0.45\,$\pm$\,0.01                                            \\
ROA~\cite{fu2023deep}                    & 0.34\,$\pm$\,0.06                                                   & 0.39\,$\pm$\,0.04                                            & 0.29\,$\pm$\,0.04                                                   & 0.35\,$\pm$\,0.03                                            \\
RWM~\cite{li2025robotic}                 & 0.90\,$\pm$\,0.03                                                   & 0.85\,$\pm$\,0.03                                            & 0.61\,$\pm$\,0.05                                                   & 0.59\,$\pm$\,0.03                                            \\ \midrule
RSSM AWARE (ours)                        & 0.91\,$\pm$\,0.00                                                   & 0.85\,$\pm$\,0.01                                            & 0.76\,$\pm$\,0.01                                                   & \textbf{0.69}\,$\pm$\,\textbf{0.01}                          \\
Transformer AWARE (ours)                 & \textbf{0.93}\,$\pm$\,\textbf{0.01}                                 & \textbf{0.87}\,$\pm$\,\textbf{0.01}                          & \textbf{0.77}\,$\pm$\,\textbf{0.01}                                 & 0.68\,$\pm$\,0.01                                            \\ \bottomrule
\end{tabular}

\end{small}
\end{table}

\section{Results and Discussion} \label{sec:results}

\textbf{Anomaly Recognition.} AWARE (Transformer backbone) achieves the highest performance in anomaly recognition (\autoref{tab:discrim-results}), yielding statistically significant improvements over the strongest baseline, RWM, for both AP and F1 across low and high noise state estimation (Appendix \autoref{tab:p-value}). Specifically, AWARE reaches 0.93 \acrshort{ap} (low noise) and 0.77 \acrshort{ap} (high noise), outperforming RWM by 3\% and 26\%, respectively. Methods relying solely on state prediction error $E_{pred}$ (RSSM, Transformer, MuJoCo) perform well under low noise but suffer an average 44\% \acrshort{ap} degradation on high noise data. Standard anomaly detection methods (Isolation Forest, Statistical Anomaly Detection) fail to reliably detect anomalies, and foundation-model baselines (AESOP, AnyAnomaly) perform 38\% worse on average than dynamics-based methods. Finally, the standard ROA latent estimator performs poorly compared to our world model prediction objective. Appendix~\ref{app:full-results} provides performance of $E_{pred}$ and $E_{lat}$ individually for each method. 

\begin{figure}[h]
    \centering
    \begin{subfigure}{0.45\textwidth}
        \includegraphics[width=\linewidth]{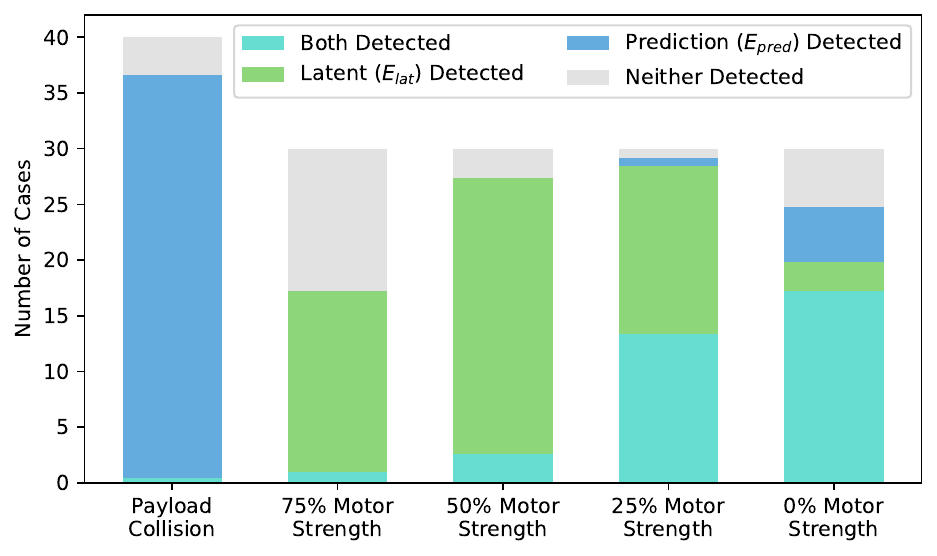}
        \caption{Low Noise State Estimation}
        \label{fig:contribution_Transformer+AWARE_paper_ground_truth}
    \end{subfigure}
    \begin{subfigure}{0.45\textwidth}
        \includegraphics[width=\linewidth]{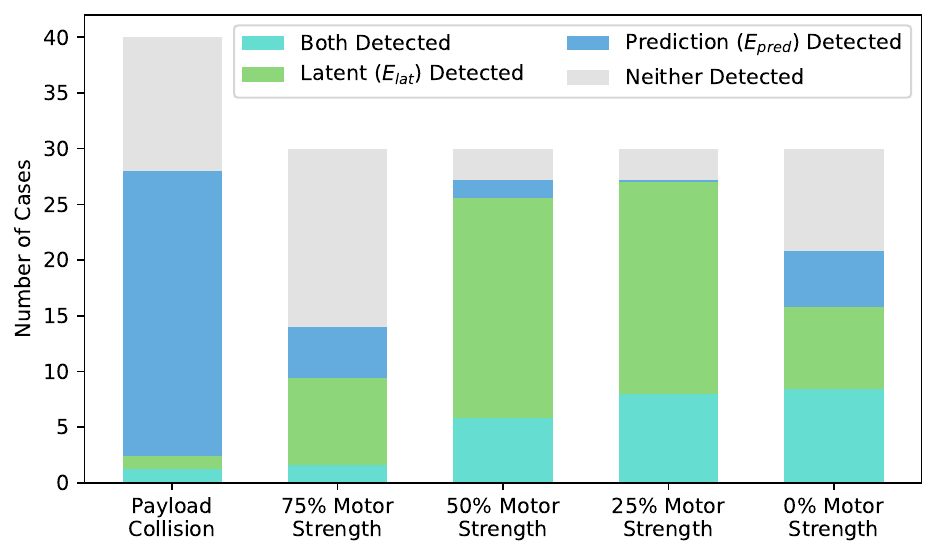}
        \caption{High Noise State Estimation}
        \label{fig:contribution_Transformer+AWARE_paper_estimated}
    \end{subfigure}
    \caption{Contribution of Latent and Predictor to Anomaly Recognition across different Impairments types using the \textit{Transformer AWARE} model under (a) \textit{Motion Capture Quality Estimation} (low noise) and (b) \textit{Camera Only Estimation} (high noise). 
    }
    \label{fig:contribution_transformer_aware}
    
\end{figure}

\autoref{fig:contribution_transformer_aware} compares the effectiveness of $E_{lat}$ and $E_{pred}$ across all the anomaly cases and shows how each of the two measurement signals contributed to solving the anomaly recognition task. $E_{lat}$ mostly detects in distribution anomalies ($75\%$, $50\%$ motor strength) and becomes less effective out-of-distribution ($0\%$, $25\%$ motor strength, payload collision). $E_{pred}$ continuously detected more cases as the amount of motor degradation increased, and was far more effective at detecting payload anomalies. These results support our hypothesis that $E_{lat}$ detects changes in modeled dynamics, $\gamma$, whilst $E_{pred}$ detects changes in the unmodeled dynamics, $\zeta$. This effect is even more significant in the low noise setting.

\begin{figure}[h]
    \centering
    \begin{subfigure}{0.45\textwidth}
        \includegraphics[width=\linewidth]{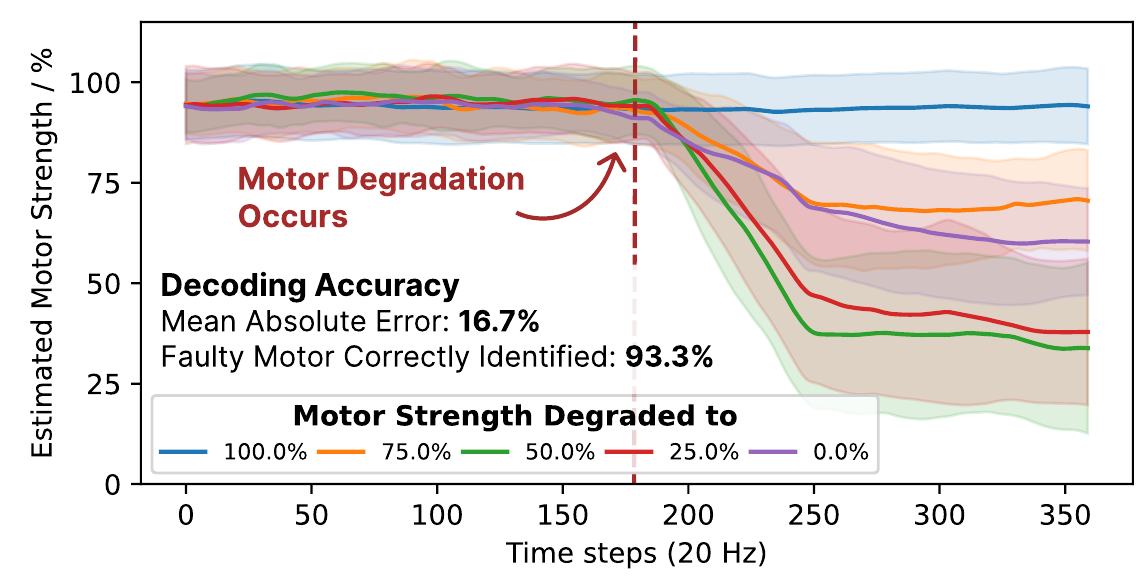}
        \caption{Low Noise State Estimation}
        \label{fig:decoder_Transformer+AWARE_paper_ground_truth}
    \end{subfigure}
    \begin{subfigure}{0.45\textwidth}
        \includegraphics[width=\linewidth]{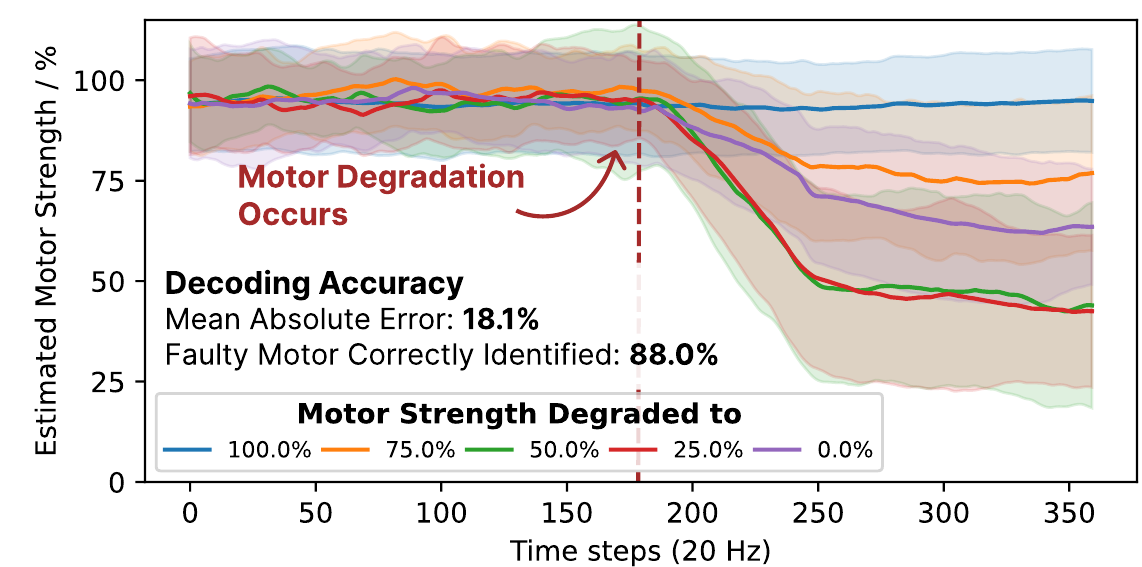}
        \caption{High Noise State Estimation}
        \label{fig:decoder_Transformer+AWARE_paper_estimated}
    \end{subfigure}
    \caption{Estimated Motor Strength using the \textit{Transformer AWARE} model under (a) Low Noise State Estimation Data and (b) High Noise State Estimation Data. The solid curves and shaded bands represent the mean and standard deviation over multiple environments respectively.  Decoding Accuracy is the rate of correctly detecting the impaired motor within a $\pm 12.5\%$ tolerance.}
    \label{fig:decoded_ctrlrange}
\end{figure}

\textbf{System Introspection. }\autoref{fig:decoded_ctrlrange} summarizes results from the introspection experiment, where we evaluate if we can recover the degraded motor strength from the estimated dynamics latent. Estimation accuracy is quantified by the mean absolute error (MAE) between the true and decoded motor strength, averaged over the final 100 timesteps of each trajectory, and aggregated across motor degradation conditions ($0\%-75\%$ of nominal strength). Under high noise estimation, the MAE is $18.1\%$, with $88.0\%$ accuracy in identifying which motor is impaired. Under low noise estimation, the MAE is $16.7\%$ and fault identification accuracy $93.3\%$.

\begin{wrapfigure}{r}{0.45\textwidth}
    \centering
    \includegraphics[width=\linewidth]{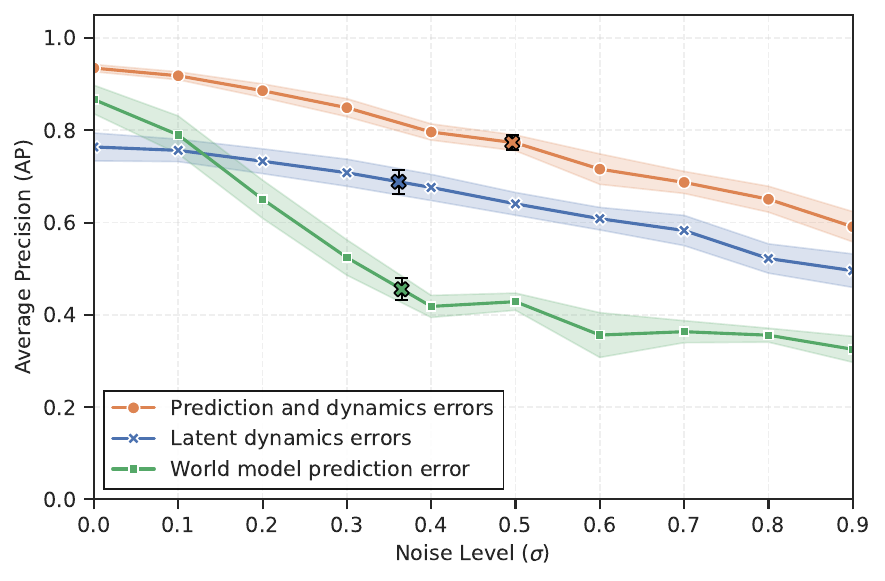}
    \caption{    
    Average Precision (AP) of Transformer AWARE across three anomaly detection modes as increasing levels of noise, sampled from a Student’s \(t\)-distribution, are injected into the Motion Capture Quality dataset. Noise is scaled relative to the standard deviation (\(\sigma\)) of each joint. The noise level of the Camera Only Quality data relative to the Motion Capture Quality data is estimated by inverse interpolation of the AP score, with the cross indicating the interpolated \(\sigma\) corresponding to the Camera Only Quality data.
    }
    \vspace{-0.4cm}
    \label{fig:AP_vs_noise}
\end{wrapfigure}

\textbf{Discussion. } AWARE outperforms all baselines in both noise conditions and are the most resilient to degrading data quality. The key factor is the dual-signal architecture ($E_{pred}, E_{lat}$). $E_{pred}$, detects unmodeled anomalies $\zeta$, and can be made more sensitive to modeled anomalies when fixed on nominal dynamics $\hat z_0$ (\autoref{tab:predictor_ablation}). However, $E_{pred}$ is inherently noise sensitive and suffers a large performance drop under camera-only state estimation (\autoref{fig:AP_vs_noise}). State prediction error only methods (RSSM, Transformer, MuJoCo) therefore see poor performance when operating in this noise setting. $E_{lat}$ is crucial for detecting modeled anomalies $\gamma$ that are missed by $E_{pred}$ (\autoref{fig:contribution_transformer_aware}). 
Additionally, $E_{lat}$ provides superior noise resiliency (\autoref{fig:AP_vs_noise}, \autoref{tab:full-discrim-results}) as it is referenced against an implicitly denoised $\hat{z}_0$. The unrestricted, world-model-informed dynamics representation learned
by AWARE demonstrates significantly better performance compared to the control-informed representation learned by ROA and raw parameter prediction from RWM. 

Foundation model based methods (AESOP and AnyAnomaly) demonstrate worse performance than dynamics-based methods across both noise levels. This confirms the importance of accurate temporal dynamics understanding over semantic generalization for robotic anomaly detection. The results for these methods are discussed further in Appendix~\ref{app: baselines}. A full discussion on the optimal way to combine $E_{pred}, E_{lat}$ and different distance measures to the reference distribution is detailed in Appendix~\ref{app:signal-processing}.

The introspection results (\autoref{fig:decoded_ctrlrange}) confirm the model learns to infer motor strength from state-action history. Estimates stabilize $\sim$80 timesteps after anomaly onset, roughly one full history window. In-distribution degradations ($75\%$, $50\%$, motor strength) are decoded more accurately than out-of-distribution cases ($25\%$, $0\%$), consistent with the limits of domain randomization.

\section{Limitations}

A core limitation of the AWARE method is its assumption of fixed operating conditions under our reference distribution. Whilst AWARE is robust to the initial distribution choice, moving to a new non-anomalous operating point (e.g. a mobile robot traversing varying terrain) may trigger false positives. Future work could address this through periodic updates of the reference distribution or defining separate intrinsic and extrinsic reference distributions to separate harmless changes in the environment from true intrinsic failures. 

We acknowledge AWARE's proven generalization is limited by our evaluation focusing on a single platform. However, AWARE is designed to make few platform specific assumptions by using domain randomization and requiring no priors on the type of anomalies that occur. Nonetheless, validation on other systems represents a valuable direction for future work. We hope that releasing our code base will help others extend this work and contribute towards safer robtic systems. 
\section{Conclusions}

We present \acrshort{aware}, a world modeling framework for robust anomaly detection under high noise state estimation. It enables passive, camera only monitoring of critical infrastructure through a novel dual-signal architecture without requiring precise system identification or any anomaly priors. On a real world robotic crane dataset of $240$ trajectories, \acrshort{aware} achieves a $93\%$ \acrshort{ap} with low-noise data, and a $77\%$ \acrshort{ap} under camera-only estimation, showing a $26\%$ improvement over baselines. \acrshort{aware} also correctly isolates the faulty motor in $88\%$ of cases. By combining state prediction with latent dynamics estimation, \acrshort{aware} bridges the gap between noise resilience and dynamic precision, providing a reliable, introspective safety layer for passive infrastructure monitoring.





\clearpage
\acknowledgments{This work is funded by the Defense Advanced Research Projects Agency (DARPA) Learning Introspective Control (LINC) program; and the Engineering and Physical Sciences Research Council [grant number UKRI5149] as well as other sponsors of BOLD (www.bold-lab.ai/sponsors).}


\bibliography{bibliography}  

\newpage
\appendix
\onecolumn



\section{AWARE Architecture}
\label{sec:details_of_aware_arch}

\subsection{Robotic Platform}

Our target platform is an under-actuated robot crane (\autoref{fig:methods}) with three actuated joints, $\mathcal{A} \in \mathbb{R}^3$, and four free joints, $\mathcal{S} \in \mathbb{R}^7$. The actuated joints are slew ($q_1$, revolute), rotating about the vertical axis of the crane base, luff ($q_2$, revolute), raising and lowering the boom, and hoist ($q_5$, prismatic), controlling cable length. The suspended payload is modeled with two 2D spherical joints at the cable endpoints: ($q_3, q_4$) at the boom captures pendulum swing and ($q_6, q_7$) at the payload captures double-pendulum effects when the payload is not aligned with the cable. 

The action $a_t$ is a velocity command for the slew, luff, and hoist joints. We simulate the system in MuJoCo and define a hidden dynamics parameter $\gamma$ to capture motor properties and payload mass: for each motor, an inertial term (\texttt{armature} in MuJoCo) and a maximum velocity term (\texttt{ctrlrange} in MuJoCo), plus payload mass, yielding $\gamma \in \mathbb{R}^7$. We focus on the motors because they dominate system dynamics and the maximum velocity term can be accurately known on the real system during experiments given privileged access. This enables the analysis of decoded motor strength in $\hat\gamma_t$ against the ground truth.

The control architecture and hardware interface are implemented within ROS 2 Humble. Actuation commands and telemetry are interfaced with the crane's control bus through a dedicated Raspberry Pi 4, allowing for necessary signal reading and writing. The perception pipeline employs two Stereolabs cameras: a ZED X\footnote{https://www.stereolabs.com/en-gb/products/zed-x} mounted off the crane and a ZED 2i\footnote{https://www.stereolabs.com/en-gb/products/zed-2i} rigidly attached to the crane structure. Both cameras interface with an NVIDIA Jetson AGX Orin, which in turn is connected to the host via Ethernet, managing state estimation and all primary processes within our ROS framework.

\subsection{Model Training and Hyperparameters} \label{app:model-hyperparameters}

The models were trained using data generated from a MuJoCo XLA simulation of the crane. To serve as a baseline and support analysis of true dynamics parameters, extensive fine-tuning of $\gamma$ was conducted using motion capture data on the real crane with a grid search over the six motor parameters. The uncovered values represent the ground truth $\gamma^*$.

We then apply uniform domain randomization centered on $\gamma^*$ with a randomized scale factor within the ranges $[\gamma_{min}, \gamma_{max}]$. The payload mass parameter was randomized absolutely. These wide ranges reflect the typical deployment case where fine-tuning is not possible, and there is high uncertainty. Table~\ref{tab:dr_params} summarizes the randomization scales across the dynamics parameters.

Training data was generated in MuJoCo using an exploratory policy which aimed to maximize coverage of the state action space. The policy would independently operate each actuated joint using one of four sub-policies: random, random-walk, sinusoid and no-motion with smoothing between action steps, using a timestep between observations of $t=0.05$~s to reflect deployment at $20$Hz. We model the real sensor noise distribution in simulation during training, combining gaussian noise with time correlated noise, and including artifacts such as repeated readings, zero error offsets, and sensor spikes. For velocity, we add noise arising from the differentiation of position. The dynamics parameters $\gamma_t$ had only gaussian noise applied, and were randomly changed during trajectories. All models were trained using a single NVIDIA L40S 48GB GPU, taking under $48$ hrs.

The models received $H=64$ timesteps of history, and made predictions over $N=20$ future timesteps during anomaly detection. The Transformer and RSSM world model backbones have 600,000 and 400,000 parameters respectively, with both latent dynamics estimators for the AWARE models having 500,000 parameters. 

\begin{table}[h]
\centering
\caption{Parameters for Domain Randomization}
\label{tab:dr_params}
\begin{tabular}{lll}
\hline
Parameter                            & Config in MuJoco & Range      \\ \hline
Motor Strength Scale Factor                      & \texttt{ctrlrange}        & (0.5, 1.5) \\
Motor Rotor Inertia Scale Factor & \texttt{armature}         & (0.1, 2.0) \\
Payload Mass / $kg$                  &                  & (0.1, 2.0) \\ \hline
\end{tabular}
\end{table}

\section{Additional Results} 

\subsection{Full Results Table} \label{app:full-results}
\begin{table}
\begin{small}
\caption{Anomaly recognition results organized by anomaly decision input. Bold entries are statistically significant improvement over the best non-AWARE model. $^*$AnyAnomaly is only evaluated on the nominal and payload collision set.}
\label{tab:full-discrim-results}
\centering

\begin{tabular}{@{}l l c c @{\hspace{1.2em}} c c@{}}
\toprule
\multirow{2}{*}{Model} &
\multirow{2}{*}{\begin{tabular}[c]{@{}l@{}}Anomaly\\ Decision\\ Input\end{tabular}} &
\multicolumn{2}{c}{\begin{tabular}[c]{@{}c@{}}Motion Capture Quality\\ State Estimation (Low Noise)\end{tabular}} &
\multicolumn{2}{c}{\begin{tabular}[c]{@{}c@{}}Camera Only State\\ Estimation (High Noise)\end{tabular}} \\
\cmidrule(lr){3-4}\cmidrule(lr){5-6}
& &
\begin{tabular}[c]{@{}l@{}}Avg.\ Precision\\$\pm$ std\end{tabular} &
\begin{tabular}[c]{@{}l@{}}F1 Score\\$\pm$ std\end{tabular} &
\begin{tabular}[c]{@{}l@{}}Avg.\ Precision\\$\pm$ std\end{tabular} &
\begin{tabular}[c]{@{}l@{}}F1 Score\\$\pm$ std\end{tabular} \\
\midrule

Randomly Generated Signal &
\multirow{4}{*}{\begin{tabular}[c]{@{}l@{}}Raw sensor\\ data\end{tabular}} &
0.17\,$\pm$\,0.03 & 0.26\,$\pm$\,0.03 &
0.17\,$\pm$\,0.03 & 0.26\,$\pm$\,0.03 \\
Isolation Forest~\cite{liu2008isolation} & &
0.29\,$\pm$\,0.00 & 0.36\,$\pm$\,0.01 &
0.16\,$\pm$\,0.00 & 0.25\,$\pm$\,0.01 \\
Statistical Anomaly Detection & &
0.40\,$\pm$\,0.01 & 0.40\,$\pm$\,0.01 &
0.22\,$\pm$\,0.01 & 0.31\,$\pm$\,0.02 \\
AESOP~\cite{SinhaElhafsiEtAl2024} & &
0.47\,$\pm$\,0.00 & 0.56\,$\pm$\,0.00 &
0.24\,$\pm$\,0.00 & 0.41\,$\pm$\,0.00 \\
AnyAnomaly$^*$~\cite{Ahn2026WACV} & &
0.37\,$\pm$\,0.07 & 0.56\,$\pm$\,0.00 &
0.37\,$\pm$\,0.07 &  0.56\,$\pm$\,0.00 \\
\midrule

MuJoCo~\cite{todorov2012mujoco} &
\multirow{6}{*}{\begin{tabular}[c]{@{}l@{}}World model\\ prediction error,\\ $E_{pred}$\end{tabular}} &
0.77\,$\pm$\,0.00 & 0.77\,$\pm$\,0.00 &
0.45\,$\pm$\,0.00 & 0.44\,$\pm$\,0.00 \\
RSSM~\cite{Hafner2025mastering} & &
0.86\,$\pm$\,0.01 & 0.80\,$\pm$\,0.01 &
0.53\,$\pm$\,0.01 & 0.56\,$\pm$\,0.01 \\
Transformer~\cite{vaswani2017attention} & &
0.84\,$\pm$\,0.01 & 0.79\,$\pm$\,0.01 &
0.41\,$\pm$\,0.02 & 0.45\,$\pm$\,0.01 \\
RWM~\cite{li2025robotic} & &
0.79\,$\pm$\,0.06 & 0.73\,$\pm$\,0.07 &
0.41\,$\pm$\,0.03 & 0.44\,$\pm$\,0.03 \\
RSSM AWARE (ours) & &
0.90\,$\pm$\,0.01 & 0.85\,$\pm$\,0.01 &
0.59\,$\pm$\,0.01 & 0.61\,$\pm$\,0.01 \\
Transformer AWARE (ours) & &
0.87\,$\pm$\,0.03 & 0.82\,$\pm$\,0.03 &
0.46\,$\pm$\,0.02 & 0.50\,$\pm$\,0.02 \\
\midrule

ROA~\cite{fu2023deep} &
\multirow{4}{*}{\begin{tabular}[c]{@{}l@{}}Latent dynamics\\ estimation error,\\ $E_{lat}$\end{tabular}} &
0.34\,$\pm$\,0.06 & 0.39\,$\pm$\,0.04 &
0.29\,$\pm$\,0.04 & 0.35\,$\pm$\,0.03 \\
RWM~\cite{li2025robotic} & &
0.75\,$\pm$\,0.04 & 0.70\,$\pm$\,0.04 &
0.53\,$\pm$\,0.05 & 0.53\,$\pm$\,0.03 \\
RSSM AWARE (ours) & &
0.70\,$\pm$\,0.02 & 0.63\,$\pm$\,0.02 &
0.61\,$\pm$\,0.02 & 0.59\,$\pm$\,0.01 \\
Transformer AWARE (ours) & &
0.76\,$\pm$\,0.03 & 0.71\,$\pm$\,0.03 &
0.69\,$\pm$\,0.03 & 0.64\,$\pm$\,0.02 \\
\midrule

RWM~\cite{li2025robotic} &
\multirow{3}{*}{\begin{tabular}[c]{@{}l@{}}Prediction and\\ dynamics errors,\\ $[E_{pred},\,E_{lat}]$\end{tabular}} &
0.90\,$\pm$\,0.03 & 0.85\,$\pm$\,0.03 &
0.61\,$\pm$\,0.05 & 0.59\,$\pm$\,0.03 \\
RSSM AWARE (ours) & &
0.91\,$\pm$\,0.00 & 0.85\,$\pm$\,0.01 &
0.76\,$\pm$\,0.01 & \textbf{0.69}\,$\pm$\,\textbf{0.01} \\
Transformer AWARE (ours) & &
\textbf{0.93}\,$\pm$\,\textbf{0.01} & \textbf{0.87}\,$\pm$\,\textbf{0.01} &
\textbf{0.77}\,$\pm$\,\textbf{0.01} & 0.68\,$\pm$\,0.01 \\
\bottomrule
\end{tabular}




\end{small}
\end{table}

Table~\ref{tab:full-discrim-results} shows the results of the anomaly detection experiment organized by the anomaly decision signal. The combination of prediction and dynamics errors, $[E_{pred}, E_{lat}]$ achieve the best results. $E_{lat}$-only methods demonstrate the best noise resilience when moving from low to high noise regimes with an average $17\%$ drop across all models. Despite this, $E_{lat}$ alone isn't sufficient for accurate anomaly detection, combining with $E_{pred}$ gives AWARE models an average $18\%$ AP improvement in the high noise setting.

\subsection{Statistical Significance Tests} \label{app:p-values}

\begin{table}
\centering
\caption{Per-metric raw $p$-values from two-sided Wilcoxon signed-rank tests over 10 paired seeds.
Significance is assessed within each baseline using a Bonferroni-corrected threshold over four metrics
(two datasets $\times$ AP/F1), i.e., $\alpha = 0.05/4 = 0.0125$.}
\label{tab:p-value}
\begin{tabular}{llcccc}
\hline
\multirow{2}{*}{} &
  \multirow{2}{*}{Baseline} &
  \multicolumn{2}{c}{\begin{tabular}[c]{@{}c@{}}Motion Capture Quality\\  (Low Noise)\end{tabular}} &
  \multicolumn{2}{c}{\begin{tabular}[c]{@{}c@{}}Camera Only\\  (High Noise)\end{tabular}} \\ \cline{3-6} 
                                                                              &                               & AP ($p$) & F1 ($p$) & AP ($p$) & F1 ($p$) \\ \hline
\multirow{11}{*}{\begin{tabular}[c]{@{}l@{}}Transformer\\ AWARE\end{tabular}} & RSSM AWARE (double)           & 0.00195  & 0.00195  & 0.02734  & 0.19336  \\
                                                                              & RWM (double)                  & 0.00195  & 0.01172  & 0.00195  & 0.00195  \\
                                                                              & Transformer AWARE (latent)    & 0.00195  & 0.00195  & 0.00195  & 0.00195  \\
                                                                              & RSSM AWARE (latent)           & 0.00195  & 0.00195  & 0.00195  & 0.00195  \\
                                                                              & RWM (latent)                  & 0.00195  & 0.00195  & 0.00195  & 0.00195  \\
                                                                              & ROA (latent)                  & 0.00195  & 0.00195  & 0.00195  & 0.00195  \\
                                                                              & Transformer AWARE (predictor) & 0.00195  & 0.00195  & 0.00195  & 0.00195  \\
                                                                              & RSSM AWARE (predictor)        & 0.00195  & 0.00586  & 0.00195  & 0.00195  \\
                                                                              & RWM (predictor)               & 0.00195  & 0.00195  & 0.00195  & 0.00195  \\
                                                                              & Transformer (predictor)       & 0.00195  & 0.00195  & 0.00195  & 0.00195  \\
                                                                              & RSSM (predictor)              & 0.00195  & 0.00195  & 0.00195  & 0.00195  \\ \hline
\multirow{11}{*}{\begin{tabular}[c]{@{}l@{}}RSSM\\ AWARE\end{tabular}}        & Transformer AWARE (double)    & 0.00195  & 0.00195  & 0.02734  & 0.19336  \\
                                                                              & RWM (double)                  & 0.01562  & 0.86523  & 0.00195  & 0.00195  \\
                                                                              & Transformer AWARE (latent)    & 0.00195  & 0.00195  & 0.00195  & 0.00195  \\
                                                                              & RSSM AWARE (latent)           & 0.00195  & 0.00195  & 0.00195  & 0.00195  \\
                                                                              & RWM (latent)                  & 0.00195  & 0.00195  & 0.00195  & 0.00195  \\
                                                                              & ROA (latent)                  & 0.00195  & 0.00195  & 0.00195  & 0.00195  \\
                                                                              & Transformer AWARE (predictor) & 0.00195  & 0.01953  & 0.00195  & 0.00195  \\
                                                                              & RSSM AWARE (predictor)        & 0.00195  & 0.01367  & 0.00195  & 0.00195  \\
                                                                              & RWM (predictor)               & 0.00195  & 0.00195  & 0.00195  & 0.00195  \\
                                                                              & Transformer (predictor)       & 0.00195  & 0.00195  & 0.00195  & 0.00195  \\
                                                                              & RSSM (predictor)              & 0.00195  & 0.00195  & 0.00195  & 0.00195  \\ \hline
\end{tabular}
\end{table}


Table~\ref{tab:p-value} reports two-sided Wilcoxon signed-rank tests over 
10 paired seeds comparing \textbf{Transformer AWARE} and \textbf{RSSM AWARE}, with every baseline across four metrics (Motion Capture Quality State Estimation / Camera Only State Estimation $\times$ AP/F1). For each baseline, we apply a Bonferroni correction across these four tests, using $\alpha=0.05/4=0.0125$. The results show strong evidence that our method outperforms the baselines: for nearly all baseline to metric pairs, the reported p-values take the minimum discrete value observed in our setting, $p=0.00195$, indicating highly consistent per-seed improvements under both low-noise and high-noise state estimation. A small number of exceptions occur for Transformer AWARE against RSSM AWARE, where we observe weaker evidence on the high-noise metrics (e.g., $p=0.02734$ and $p=0.1934$), suggesting less consistent improvements under Camera Only State Estimation for RSSM AWARE. Similarly, for RSSM AWARE against RWM, we also observe weaker evidence on the F1 score metric in the low-noise setting ($p=0.86523$). 

\subsection{World Model Prediction Error} \label{app:prediction-error}
\begin{figure}
    \centering
    \includegraphics[width=0.75\linewidth]{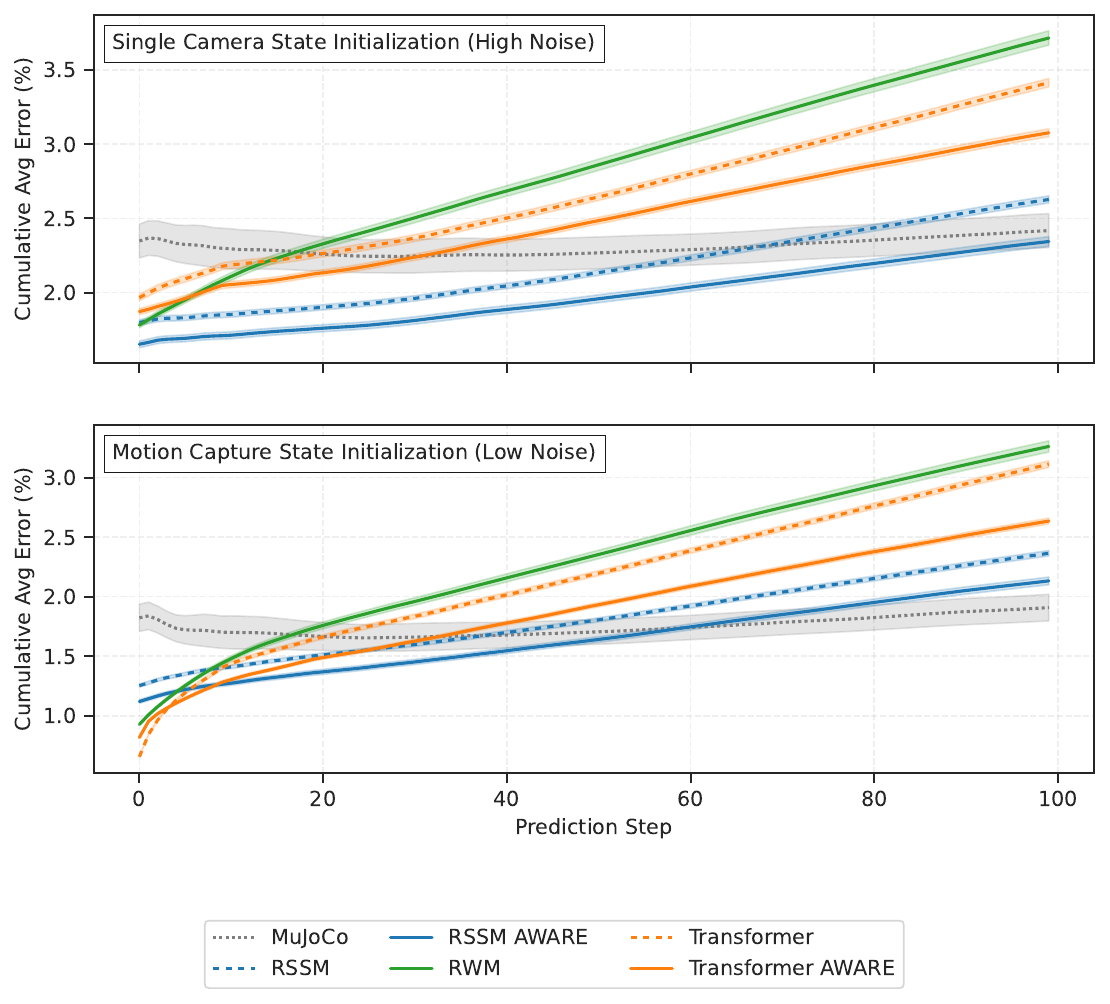}
    \caption{World model prediction error over a 100 step horizon. We condition the model on either Single Camera Quality Data or Motion Capture Quality Data, computing the error relative the Motion Capture data in both cases. The error is computed as a fraction of the joint's total range, averaged over all joint dimensions then as a running cumulative error averaged over all experiments.} 
    \label{fig:wm-prediction-error}
\end{figure}

We investigate the world model prediction accuracy by comparing foretasted predictions to the real world crane over a 100 step horizon, equivalent to 5 seconds, in \autoref{fig:wm-prediction-error}. We test initializing the model on both low and high noise data, with the errors being calculated in both cases using the low noise motion capture data of the crane. We find the AWARE enhanced world models perform better than their non-AWARE counterpart, demonstrating the impact of the encoded dynamics parameter latent in making accurate predictions. We additionally see that all the learned world models are more accurate than MuJoCo over a short horizon, showing our domain randomization procedure helps effectively cross the sim-to-real gap. The RSSM AWARE model is the strongest learned predictor and when conditioned on noisy data is better than MuJoCo over the full 100 step horizon, likely due to it's recurrent hidden state architecture favoring dynamics prediction, as seen in prior works \cite{hansen2024tdmpc, hafner2019pixels}.

\section{Ablation Studies}

\begin{figure}[h]
    \centering
    \includegraphics[width=1.0\linewidth]{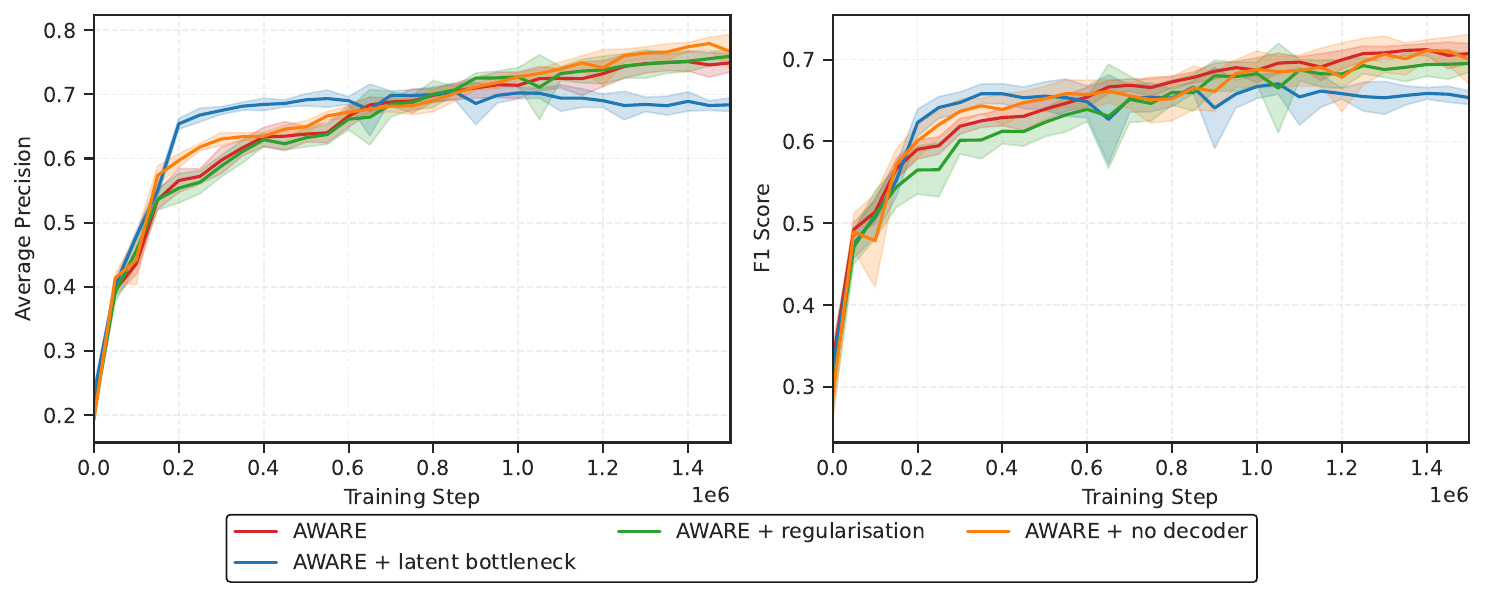}
    \caption{Training curves for RSSM AWARE. We ablate by reintroducing components from \textit{Regularized Online Adaptation} \cite{fu2023deep} to AWARE, including: regularization of the target encoder that the latent dynamics estimator learns to follow; removing the latent decoder and reintroducing the latent bottleneck. We plot both average precision and F1 over 1.5 million training steps on the Single Camera Quality dataset.} 
    \label{fig:ablation_study}
\end{figure}

\subsection{AWARE Components}
We ablate over the changes we make to the Real Time Online Adaptation method \cite{fu2023deep} for learning an latent dynamics estimator. We reintroduced each of three core changes we made from the original method: regularizing the encoder, removing the decoder and reintroducing a latent bottleneck. \autoref{fig:ablation_study} shows the training curves across the ablation over 1.5 training steps for the RSSM. We see a key improvement in AWARE is removing the latent bottleneck, suggesting a large latent space is important for encoding information useful for dynamics prediction. Encoder regularization, which is designed to improve the performance of the student latent dynamics estimator has a negligible effect. We argue removing it is therefore beneficial as it simplifies the method, requiring fewer parameters to tune. Removing the decoder sees little performance gain, coming at the cost of introspection.

\subsection{Prediction Anomaly Score}\label{subsec:prediction-ablation}
We investigate the method discussed in \autoref{subsec:anomaly-decision} to use the world model's predictions to detect anomalies by seeing if the model's state predictions vary from the observed predictions, aiming to detect changes in the unmodeled dynamics $\zeta$. \autoref{tab:predictor_ablation} shows these results for the RSSM AWARE method. We first test the value of encoding the dynamics parameters to a latent representation $\hat{z}$, as opposed to directly using parameters $\gamma$ as in Robotic World Model \cite{li2025robotic}. We observed a significant performance degradation in anomaly detection when not encoding the parameters, suggesting we are able to learn a more useful latent representation of the parameters that enables better quality prediction. We then test not normalizing the model error by the variance predicted by the model (\autoref{equation:anomaly-pred}). We hypothesized that this would reduce the contribution of errors to the anomaly score when the model is uncertain. We see marginal reduction under low noise data but more significant reduction under high noise data when the model is more likely to be uncertain given it has been conditioned on noise which supports our hypothesis. Finally, we test not conditioning the model on fixed median nominal latent $\hat{z_0}$, instead conditioning the model on the moving latent representation of the dynamics parameters $\hat{z}_t$. This significantly reduces performance, suggesting our model is better able to capture anomalies when solely conditioned on the nominal case, as opposed to using a potential anomalous latent. 

\begin{table}[ht]
\centering
\caption{RSSM AWARE results using only the world model $E_{pred}$ to detect anomalies. We ablate over the enhancements we make to improve predictor performance, including learning a latent representation, $\hat{z}$, of the hidden dynamics parameter, $\gamma$; conditioning the world model on the fixed median nominal latent, $\hat{z}_0$; and normalizing the computed error by the world model's variance.}
\label{tab:predictor_ablation}
\small
\begin{tabularx}{\textwidth}{@{} >{\raggedright\arraybackslash}X >{\raggedright\arraybackslash}X >{\raggedright\arraybackslash}X c c c c @{}}
\toprule
\multirow{3}{*}{\begin{tabular}[c]{@{}l@{}}Encode \\ dynamics \\ param to $\hat{z}_t$\end{tabular}} &
  \multirow{3}{*}{\begin{tabular}[c]{@{}l@{}}Dynamics \\ latents fixed \\ to median $\hat{z}_0$\end{tabular}} &
  \multirow{3}{*}{\begin{tabular}[c]{@{}l@{}}Error \\ normalized by \\ variance $\sigma_t$\end{tabular}} &
  \multicolumn{2}{c}{Motion Capture Quality} &
  \multicolumn{2}{c}{Single Camera Quality} \\
           &            &            & \multicolumn{2}{c}{State Estimation (Low Noise)} & \multicolumn{2}{c}{State Estimation (High Noise)} \\ \cmidrule(l){4-7} 
           &            &            & AP                      & F1                     & AP                      & F1                      \\ \midrule
& \checkmark & \checkmark & 0.71 $\pm$ 0.02 & 0.65 $\pm$ 0.02 & 0.50 $\pm$ 0.01 & 0.54 $\pm$ 0.01 \\
\checkmark & \checkmark & & 0.88 $\pm$ 0.01 & 0.83 $\pm$ 0.01 & 0.55 $\pm$ 0.01 & 0.58 $\pm$ 0.01 \\
\checkmark & & \checkmark & 0.81 $\pm$ 0.01 & 0.75 $\pm$ 0.01 & 0.50 $\pm$ 0.01 & 0.54 $\pm$ 0.01 \\
\checkmark & \checkmark & \checkmark & 0.90 $\pm$ 0.01 & 0.85 $\pm$ 0.01 & 0.59 $\pm$ 0.01 & 0.61 $\pm$ 0.01 \\
\bottomrule
\end{tabularx}
\end{table}

\subsection{Foundation Models}
\label{app: baselines}

Foundation models are promising for anomaly detection under passive state estimation as they aim to provide a semantic level understanding of the scene and thus not requiring a noisy estimate of the robot state. We test two methods: LLM based AESOP ~\cite{SinhaElhafsiEtAl2024} and VLM based AnyAnomaly ~\cite{Ahn2026WACV} on our robotic crane dataset. 

\textbf{AESOP. }AESOP use a small LLM conditioned on a prompt that describes the task overview, observations ~\cite{SinhaElhafsiEtAl2024}. For our dataset, we use the estimated joint states and actions as observation inputs that provide dynamics context of the scene. The language model then outputs an latent embedding which is used to compare with a precomputed reference via a same Mahalanobis distance measurement. In our experiments, we note that the model is struggling to encode differentiated latents from raw observations, as it provides limited dynamic semantics. To resolve this, we translate observations into qualitative text descriptions before filling the prompt templates. Joint velocities are mapped into five categorical linguistic bins("very slow", "slow", "moderate", "fast", and "very fast"), defined by the percentile ranges computed across the nominal training dataset. For double pendulum joints ($q_3, q_4, q_6, q_7$), we compute the L2 norm and then use it for categorization. We provide an example prompt as follows:

\begin{promptbox}[Example Prompt]
\label{promptbox:AESOP}
\small\ttfamily
Task: A crane is transporting a payload to the target drop zone. \\
Observations: slew: commanded hard left, slewing left slowly; luff: commanded down, luffing down slowly; hoist: no hoist command, hoist stationary; boom tip swinging slowly; payload stable.
\end{promptbox}

The anomaly detection results are compared in Table~\ref{tab:discrim-results}. To explain why AESOP underperforms (-46\%) relative to AWARE, we propose a two-fold argument. Firstly, AESOP's capacity is capped by the use of a lightweight language model (MPNet-110M) mandated by real-time inference constraints. Crucially, however, the model encodes textual descriptions rather than visual data. The reliance on a reduced representation results in a substantial loss of the rich semantics inherent in the original raw images.

\textbf{AnyAnomaly. }AnyAnomaly is a zero-shot anomaly detection method that uses a VLM and guided prompt to determine if an anomaly specified in the prompt is occurring in the given batch of frames. The method operates temporally over a short sequence of 4 key frames. Given this limited window, evaluating the method on our motor degradation dataset, which requires long-horizon dynamics understanding, was out of the intended scope of this work. We therefore only choose to apply our payload collision dataset alongside the nominal cases. We utilize the official implementation and prompt template, filling in with our specified task: 

\begin{promptbox}[Example Prompt]
\label{promptbox:AnyAnomaly}
\small\ttfamily
Task: Evaluate whether the given image includes \textit{a person manipulating the red payload or the invisible string that it is vertically suspended from} on a scale from 0 to 1.
\end{promptbox}

Using Qwen3-VL-8B-Instruct \cite{qwen3technicalreport} as the VLM backbone, AnyAnomaly achieves AP of 0.37 and F1 of 0.56. Whilst being the only model that is designed for external camera state estimation, it struggled on our dataset with a high false negative rate ($21/40$). We found the VLM paradigm faced two core challenges in this domain: spatial misidentification resulting in background objects being identified as the crane and a lack of precise physical scene understanding, resulting in some subtle visible manipulations not to be flagged by the model. Furthermore, we found performance to be highly sensitive to prompt phrasing, requiring multiple iterations to achieve reasonable results. These limitations highlight the motivation behind AWARE. Explicitly learning the temporal dynamics enables higher reliability and sensitivity to anomalies than semantic visual interpretation.

\clearpage
\section{Anomaly Recognition}
\subsection{Measurement Signal Processing} \label{app:signal-processing}

We calculate the anomaly measurement by computing the Mahalanobis distance of the prediction error $E_{i,t}$ against reference distribution $\mathcal{P}_{ref}=(\mu_i, \sum_i)_{ref}$:
\begin{equation}
    M_{i,t}=\sqrt{\left(E_{i,t}-\mu_{ref}\right)^T\sum^{-1}\left(E_{i,t}-\mu_{ref}\right)}
\end{equation}
where $i=lat, pred$.

The Mahalanobis distance is appealing, being statistically justified, versatile, and simple to implement. It provided the better performance even than hand-tuned normalization, likely due to accounting for the full covariance matrix. As shown in \autoref{fig:AP_vs_noise} and \autoref{tab:discrim-results}, the latent signal is general more stable and more resilient to noise that the world model prediction error, as it takes a longer term view and aims to uncover the pattern rather than analyzing short horizon behavior. It therefore follows that long term degradations, such as motor impairment, are well detected by the latent signal, whereas, short term instantaneous events, such as the payload collisions, are best detected by the world model error.

\autoref{tab:distance-ablation} justifies our decision to use the Mahalanobis distance. We test multiple distance measures to our reference distribution and different operators to combine our two anomaly signals $E_{pred}$ and $E_{lat}$. Notably, we see significant performance increase when moving from L1/L2 measurements to L2 standardized and Mahalanbois measures. The demonstrates the importance of normalizing by the diagonal of the covariance matrix and reduces noise from scale invariance between the features. We see a marginal improvement moving from L2 standardized to Mahalanbois, suggesting the impact of using the full covariance matrix (and therefore accounting for feature correlations) is small.

\begin{table}[b]
\centering
\caption{Anomaly detection results averaged across AWARE methods using different distance measures to the nominal reference distribution and operators to combine the prediction and latent signal $E_{pred}$ and $E_{lat}$. 'L1' and 'L2' use the Manhattan and Euclidean distance to the reference distribution mean. 'L2 standardized' normalizes the L2 distance by the diagonal covariance.}
\label{tab:distance-ablation}

\begin{tabular}{@{}l l c c @{\hspace{1.2em}} c c@{}}
\toprule
\multirow{2}{*}{\begin{tabular}[c]{@{}l@{}}Distance\\ Measure\end{tabular}} &
\multirow{2}{*}{Operator} &
\multicolumn{2}{c}{\begin{tabular}[c]{@{}c@{}}Motion Capture Quality\\ State Estimation (Low Noise)\end{tabular}} &
\multicolumn{2}{c}{\begin{tabular}[c]{@{}c@{}}Camera Only State\\ Estimation (High Noise)\end{tabular}} \\
\cmidrule(lr){3-4}\cmidrule(lr){5-6}
& &
\begin{tabular}[c]{@{}l@{}}Avg.\ Precision\\$\pm$ std\end{tabular} &
\begin{tabular}[c]{@{}l@{}}F1 Score\\$\pm$ std\end{tabular} &
\begin{tabular}[c]{@{}l@{}}Avg.\ Precision\\$\pm$ std\end{tabular} &
\begin{tabular}[c]{@{}l@{}}F1 Score\\$\pm$ std\end{tabular} \\
\midrule

\multirow{3}{*}{L1} &
OR  & 0.86 $\pm$ 0.02 & 0.80 $\pm$ 0.02 & 0.69 $\pm$ 0.03 & 0.63 $\pm$ 0.04 \\
& AND & 0.78 $\pm$ 0.04 & 0.72 $\pm$ 0.03 & 0.66 $\pm$ 0.01 & 0.59 $\pm$ 0.04 \\
& SUM & 0.78 $\pm$ 0.09 & 0.72 $\pm$ 0.07 & 0.50 $\pm$ 0.04 & 0.55 $\pm$ 0.04 \\
\midrule

\multirow{3}{*}{L2} &
OR  & 0.83 $\pm$ 0.03 & 0.78 $\pm$ 0.02 & 0.67 $\pm$ 0.04 & 0.60 $\pm$ 0.04 \\
& AND & 0.75 $\pm$ 0.05 & 0.71 $\pm$ 0.03 & 0.65 $\pm$ 0.02 & 0.59 $\pm$ 0.04 \\
& SUM & 0.64 $\pm$ 0.18 & 0.61 $\pm$ 0.14 & 0.42 $\pm$ 0.04 & 0.49 $\pm$ 0.06 \\
\midrule

\multirow{3}{*}{L2 standardized} &
OR  & 0.92 $\pm$ 0.02 & 0.86 $\pm$ 0.03 & 0.75 $\pm$ 0.03 & 0.67 $\pm$ 0.02 \\
& AND & 0.88 $\pm$ 0.02 & 0.83 $\pm$ 0.02 & 0.69 $\pm$ 0.02 & 0.61 $\pm$ 0.03 \\
& SUM & 0.91 $\pm$ 0.01 & 0.86 $\pm$ 0.02 & 0.63 $\pm$ 0.03 & 0.66 $\pm$ 0.03 \\
\midrule

\multirow{3}{*}{Mahalanobis} &
OR  & \textbf{0.92 $\pm$ 0.01} & 0.86 $\pm$ 0.02 & \textbf{0.77 $\pm$ 0.02} & \textbf{0.68 $\pm$ 0.01} \\
& AND & 0.88 $\pm$ 0.02 & 0.83 $\pm$ 0.03 & 0.70 $\pm$ 0.02 & 0.62 $\pm$ 0.03 \\
& SUM & 0.92 $\pm$ 0.01 & \textbf{0.87 $\pm$ 0.01} & 0.66 $\pm$ 0.02 & 0.67 $\pm$ 0.02 \\
\bottomrule
\end{tabular}

\end{table}

\autoref{fig:contribution_transformer_aware} compares the effectiveness of $E_{lat}$ and $E_{pred}$ across all the anomaly cases and shows how each of the two measurement signals in the AWARE model, from the world model error (predictor) and the latent estimation (latent), contributed to solving the anomaly recognition task. It illustrates that $E_{lat}$ became less reliable out-of-distribution. The number of cases correctly detected by $E_{lat}$ as motor strength reduces from $25\% \rightarrow 0\%$ decreases, even though $0\%$ represents the more significant anomaly. In contrast, $E_{pred}$ continuously detected more cases as the amount of motor degradation increased, and was far more effective at detecting payload anomalies. These results support our hypothesis that $E_{lat}$ detects changes in modeled dynamics, $\gamma$, whilst $E_{pred}$ detects changes in the unmodeled dynamics, $\zeta$. This effect is even more significant in the low noise setting.

\begin{figure}
    \centering
    \begin{subfigure}{0.95\textwidth}
        \includegraphics[width=\linewidth]{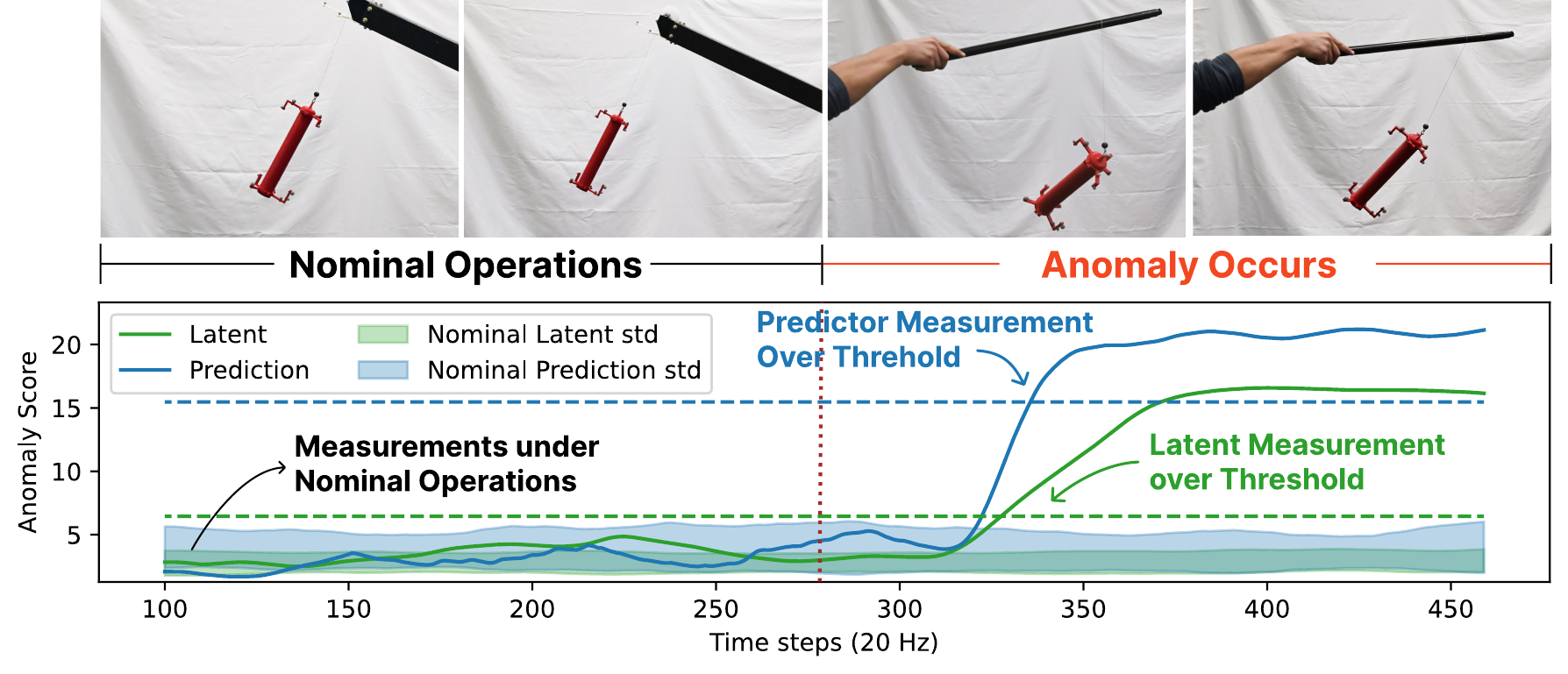}
        \caption{Both Caught}
        \label{fig:measurement_both_caught}
    \end{subfigure}
    \begin{subfigure}{0.95\textwidth}
        \includegraphics[width=\linewidth]{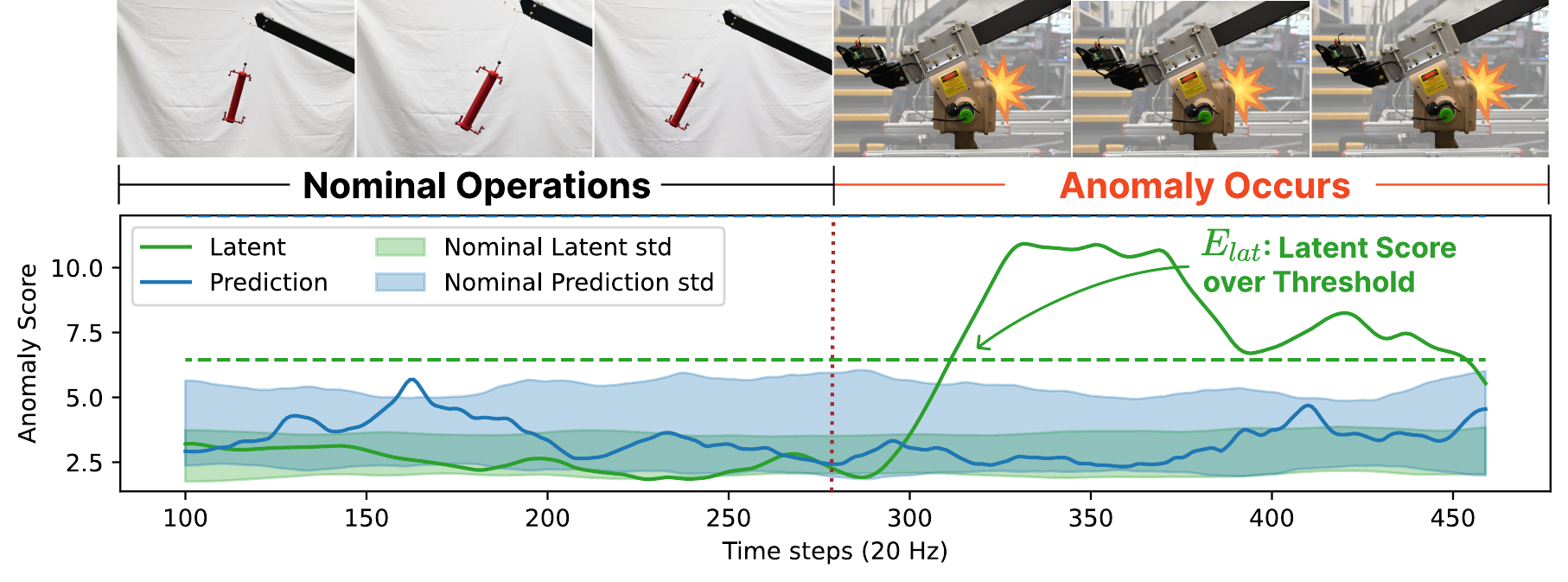}
        \caption{Latent Caught}
        \label{fig:measurement_latent_caught}
    \end{subfigure}
    \begin{subfigure}{0.95\textwidth}
        \includegraphics[width=\linewidth]{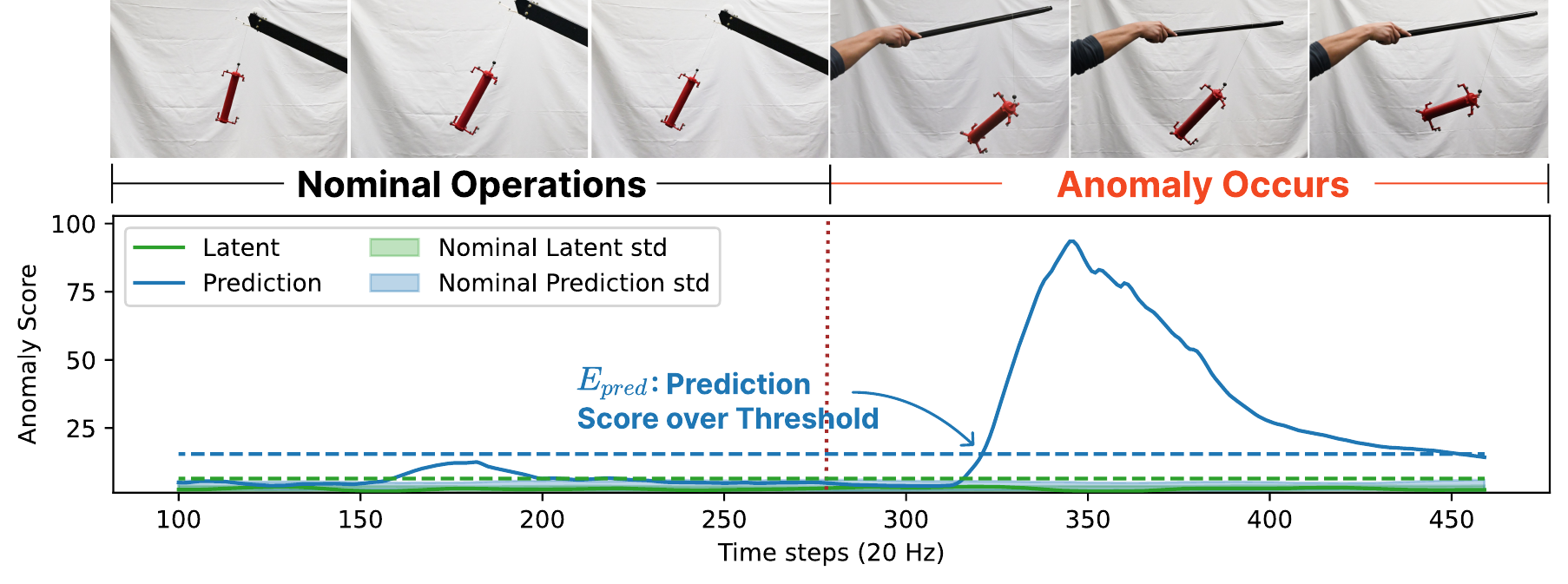}
        \caption{Predictor Caught}
         \label{fig:measurement_predictor_caught}
    \end{subfigure}
    \caption{Examples of Anomaly Recognition by a pivoting measurement signal, where a successful detection is classified as either measurement signal exceeds its corresponding threshold. Nominal std gives the standard deviation of both measurement signals during nominal operations across all environments.}
    \label{fig:measurement_illustration}
\end{figure}

\begin{figure}[p]
    \centering
    \begin{subfigure}{\textwidth}
        \includegraphics[width=\linewidth]{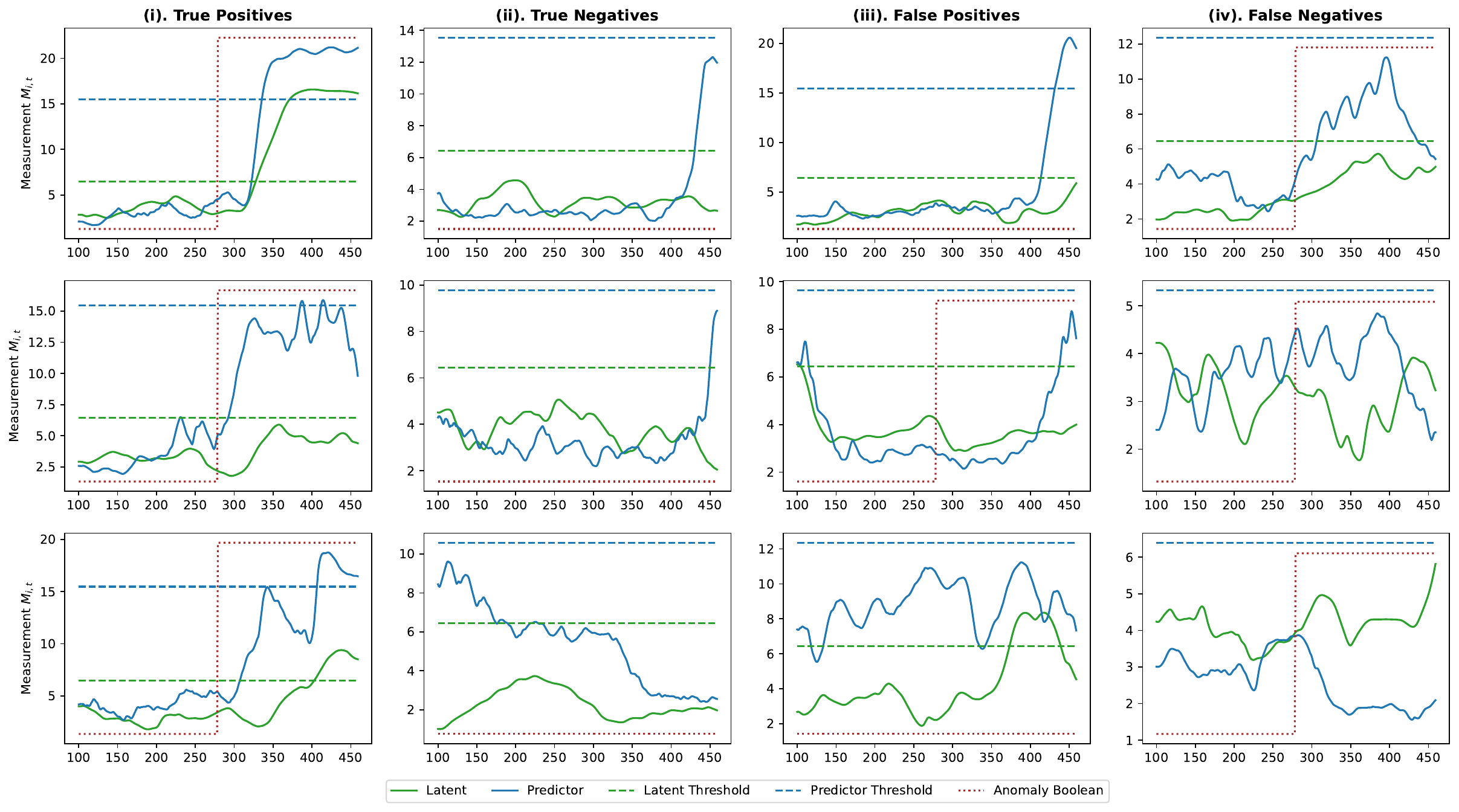}
        \caption{Motion Capture Quality Estimation}
        \label{fig:discrim_measurements_Transformer+AWARE_paper_ground_truth}
    \end{subfigure}
    \begin{subfigure}{\textwidth}
        \includegraphics[width=\linewidth]{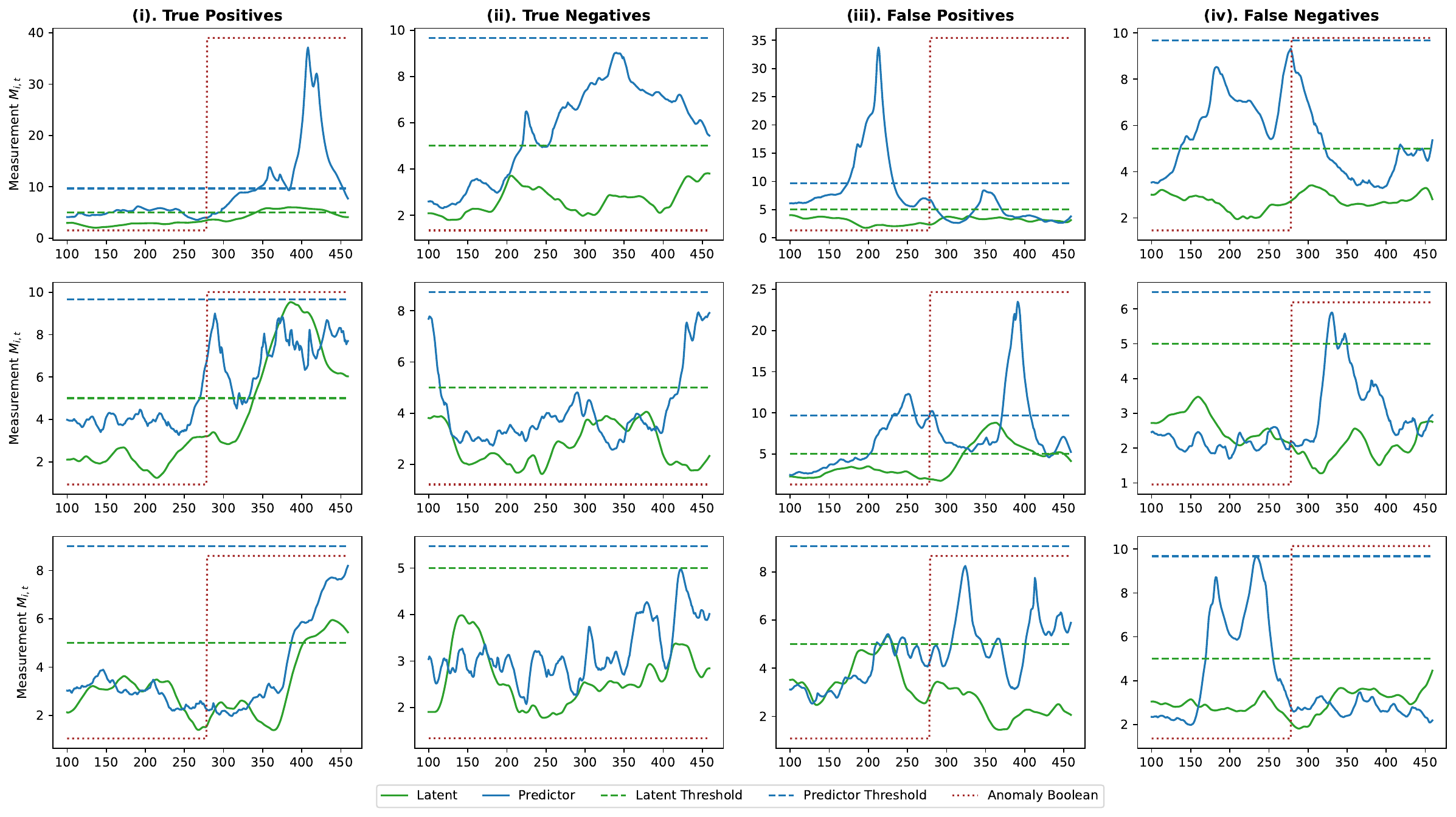}
        \caption{Camera Only Estimation}
        \label{fig:discrim_measurements_Transformer+AWARE_paper_estimated}
    \end{subfigure}
    \caption{Examples of predictor and latent-estimator measurement signals with and without anomalies. (a) \textit{Motion Capture Quality Estimation} (low noise); (b) \textit{Camera Only Estimation} (high noise). From left to right, the four columns correspond to True Positive, True Negative, False Positive, and False Negative cases (based on whether an anomaly was present and whether the model raised an alarm). Solid blue and green curves show the predictor and latent-estimator signals, respectively, while dashed blue and green lines denote their anomaly thresholds. The red dashed vertical line marks the time at which the anomaly is injected.}
    \label{fig:discrim_measurements_transformer_aware}
\end{figure}

\end{document}